%% file: acl_latex.tex
\documentclass[11pt]{article}

\usepackage[preprint]{acl}

\usepackage{times}
\usepackage{latexsym}

\usepackage[T1]{fontenc}
\usepackage[utf8]{inputenc}

\usepackage{microtype}

\usepackage{inconsolata}

\usepackage{graphicx}

\usepackage{amsmath}
\usepackage{xspace}
\usepackage{amsfonts}
\usepackage{amssymb}
\usepackage{algorithm}
\usepackage{algorithmic}
\usepackage[most]{tcolorbox}

\usepackage{booktabs}
\usepackage{multirow}
\usepackage{xcolor}
\usepackage{colortbl}
\usepackage{array}
\usepackage{makecell}
\usepackage{siunitx}
\usepackage{subcaption}
\usepackage{pifont}
\usepackage{fontawesome5}
 
\definecolor{headerblue}{RGB}{13, 71, 161}
\definecolor{groupgray}{RGB}{240, 242, 245}
\definecolor{bestgold}{RGB}{255, 248, 220}
\definecolor{oursblue}{RGB}{232, 240, 254}

\definecolor{PromptBlue}{HTML}{EAF2FF}
\definecolor{PromptFrame}{HTML}{3B6EA8}
\definecolor{PromptTitle}{HTML}{1F4E79}

\newcommand{\cmark}{\ding{51}}
\newcommand{\xmark}{\ding{55}}

\newcommand{\logo}{\raisebox{-0.8em}{\includegraphics[height=2.5em]{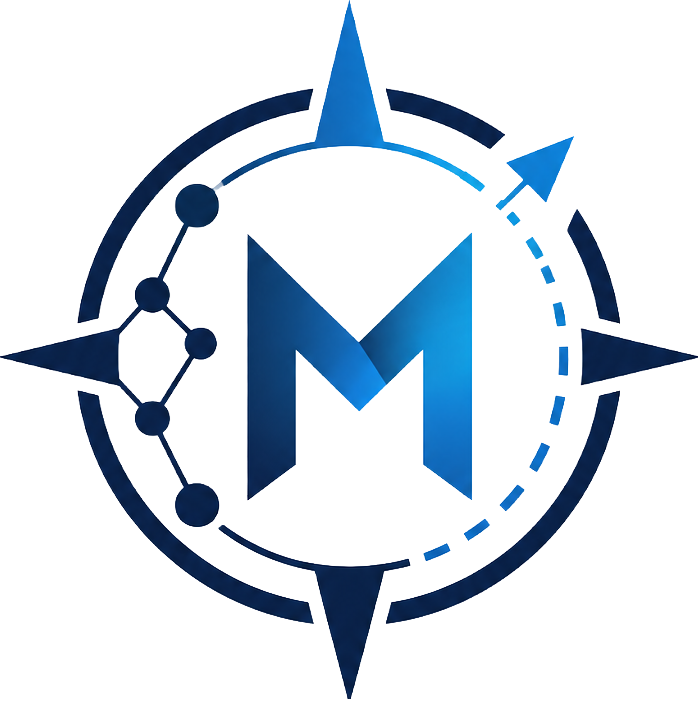}}\hspace{0.3em}}
\newcommand{\name}[1]{\textsc{MCompassRAG}\xspace}

\title{\logo \name{}: Topic Metadata as a Semantic Compass for Paragraph-Level Retrieval}

\author{
\textbf{
Amirhossein Abaskohi$^{1}$\thanks{Corresponding author: \texttt{aabaskoh@cs.ubc.ca}},
Raymond Li$^{1}$,
Gaetano Cimino$^{2}$,}\\
\textbf{
Peter West$^{1}$,
Giuseppe Carenini$^{1}$,
Issam H. Laradji$^{1,3}$}\\[0.5em]
$^{1}$University of British Columbia, $^{2}$University of Salerno, $^{3}$ServiceNow Research 
}

\newtcolorbox[auto counter, number within=section]{promptbox}[2][]{
    enhanced,
    breakable,
    colback=PromptBlue,
    colframe=PromptFrame,
    coltitle=white,
    colbacktitle=PromptTitle,
    title={\faMagic\hspace{0.5em}Prompt~\thetcbcounter: #2},
    fonttitle=\bfseries\small,
    fontupper=\small,
    boxrule=0.7pt,
    arc=2mm,
    left=7pt,
    right=7pt,
    top=6pt,
    bottom=6pt,
    attach boxed title to top left={xshift=2mm,yshift=-2mm},
    boxed title style={
        arc=2mm,
        boxrule=0pt,
        left=5pt,
        right=5pt,
        top=3pt,
        bottom=3pt
    },
    #1
}

\begin{document}
\maketitle

% -----------------------------------------------
% SECTIONS
% -----------------------------------------------

\input{sections/1_abstract}
\input{sections/2_introduction}
\input{sections/3_relatedwork}
\input{sections/4_method}
\input{sections/5_experiments}
\input{sections/6_ablations}
\input{sections/7_conclusion}

% -----------------------------------------------
% Mandatory Sections
% -----------------------------------------------

\section*{Limitations}
\name{} has a few limitations worth noting. First, the quality of topic-guided retrieval is directly dependent on the quality of the underlying topic model: poorly trained or misaligned topic representations will produce uninformative metadata signals. This creates a dependency on reliable topic modeling, which can be difficult in low-resource or specialized domains. Second, \name{} introduces several hyperparameters, including the number of topic-model topics $K$, selected metadata entries from the memory bank $L$, metadata topics used for retrieval $M$, and retrieved chunks $k$, whose interactions are non-trivial to tune. As shown in Section~\ref{sec:ablations}, performance is sensitive to the number of topics, so this choice requires validation. Third, the current topic enrichment strategy represents each chunk and query as a weighted sum of topic centroid embeddings, which is a lossy compression: combining multiple topic vectors into a single aggregated vector discards the individual structure of each topic signal. As more topics are included, aggregation becomes noisier. Future work should explore efficient sparse or cross-attention topic integration that better preserves per-topic structure.

% -----------------------------------------------
% References
% -----------------------------------------------

\bibliography{custom}

% -----------------------------------------------
% Appendix
% -----------------------------------------------

\clearpage

\appendix

\input{sections/app_prompt}
\input{sections/app_benchmarks_baseline_detailed}
\input{sections/app_training_and_implementation_details}
\input{sections/app_detailed_results}
\input{sections/app_topic_granularity}
\input{sections/app_in_domain_topic_model}
\input{sections/app_qualitative_analysis}

\end{document}

%% file: sections/1_abstract.tex
\begin{abstract}
Retrieval-augmented generation (RAG) systems depend critically on how documents are chunked and searched. Fine-grained chunks can improve retrieval precision but expand the search space, increasing latency and cost; larger chunks reduce the number of candidates but make dense similarity less reliable, as the representation for each chunk mixes multiple topics and introduces more semantic noise. This trade-off becomes especially limiting in deep research tasks, where retrieval must be both fast and precise across large, heterogeneous corpora. We introduce \textbf{\name{}}, a metadata-guided retrieval framework that uses topic-level signals as a semantic compass for selecting relevant evidence. Instead of relying only on cosine similarity between queries and noisy chunk embeddings, \name{} enriches chunk representations with topic metadata in the same embedding space and trains a lightweight retriever through LLM-teacher distillation. At inference time, \name{} performs topic-aware retrieval without additional LLM calls, improving both efficiency and evidence quality. Across six complex retrieval benchmarks, \name{} improves information efficiency (IE) by \textbf{8.24\%} on average with over \textbf{5$\times$ lower latency} than the strongest efficient RAG baselines\footnote{
Code is available on \href{https://github.com/AmirAbaskohi/MCompassRAG}
{{\textcolor{black}{\faGithub}}~GitHub}.
}. 
\end{abstract}

%% file: sections/2_introduction.tex
\section{Introduction}

Retrieval-augmented generation (RAG) has become a standard paradigm for grounding large language models (LLMs) in external knowledge~\cite{lewis2020retrieval, karpukhin-etal-2020-dense}. Yet the efficiency and quality of RAG hinge on a simple but consequential design choice: how documents are divided into retrievable units. This choice becomes especially important in deep research tasks~\cite{zhang2025deepresearchsurveyautonomous}, where systems must search large corpora and often issue many retrieval calls before producing a final answer. Standard dense retrieval over fixed-size chunks~\cite{10.1145/3637870} faces a granularity trade-off. Fine-grained chunks, such as sentences or short paragraphs, offer precise evidence but greatly increase the number of candidates to index and search. Larger chunks reduce the search space and improve retrieval efficiency, but they mix multiple topics and discourse roles into a single embedding. As a result, similarity scores become noisy: relevant evidence can be diluted by unrelated text, while partially relevant chunks may be retrieved despite containing mostly irrelevant content.

\begin{figure*}[t]
    \centering
    \begin{subfigure}[t]{0.60\textwidth}
        \centering
        \includegraphics[width=\linewidth]{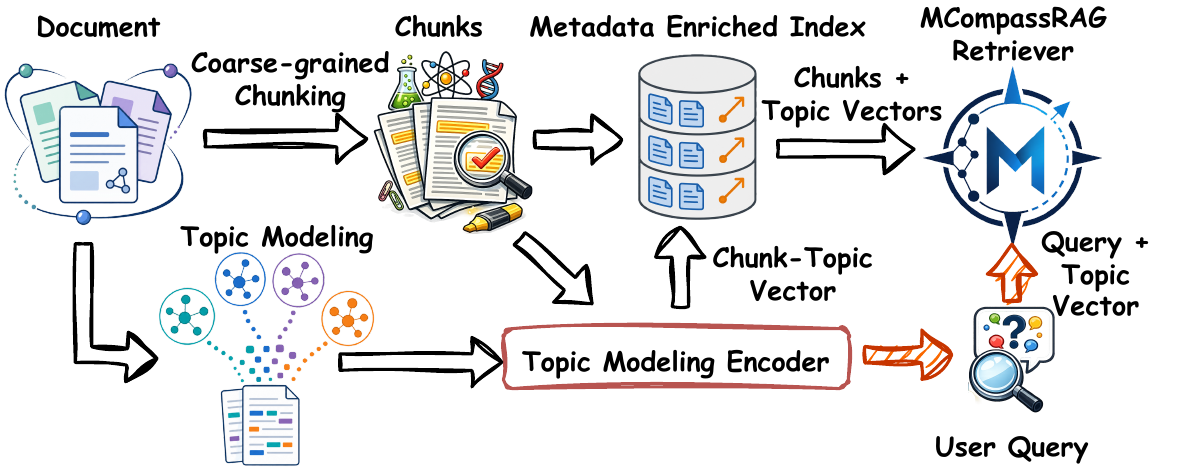}
        \caption{}
        \label{fig:teaser}
    \end{subfigure}
    \begin{subfigure}[t]{0.36\textwidth}
        \centering
        \includegraphics[width=\linewidth]{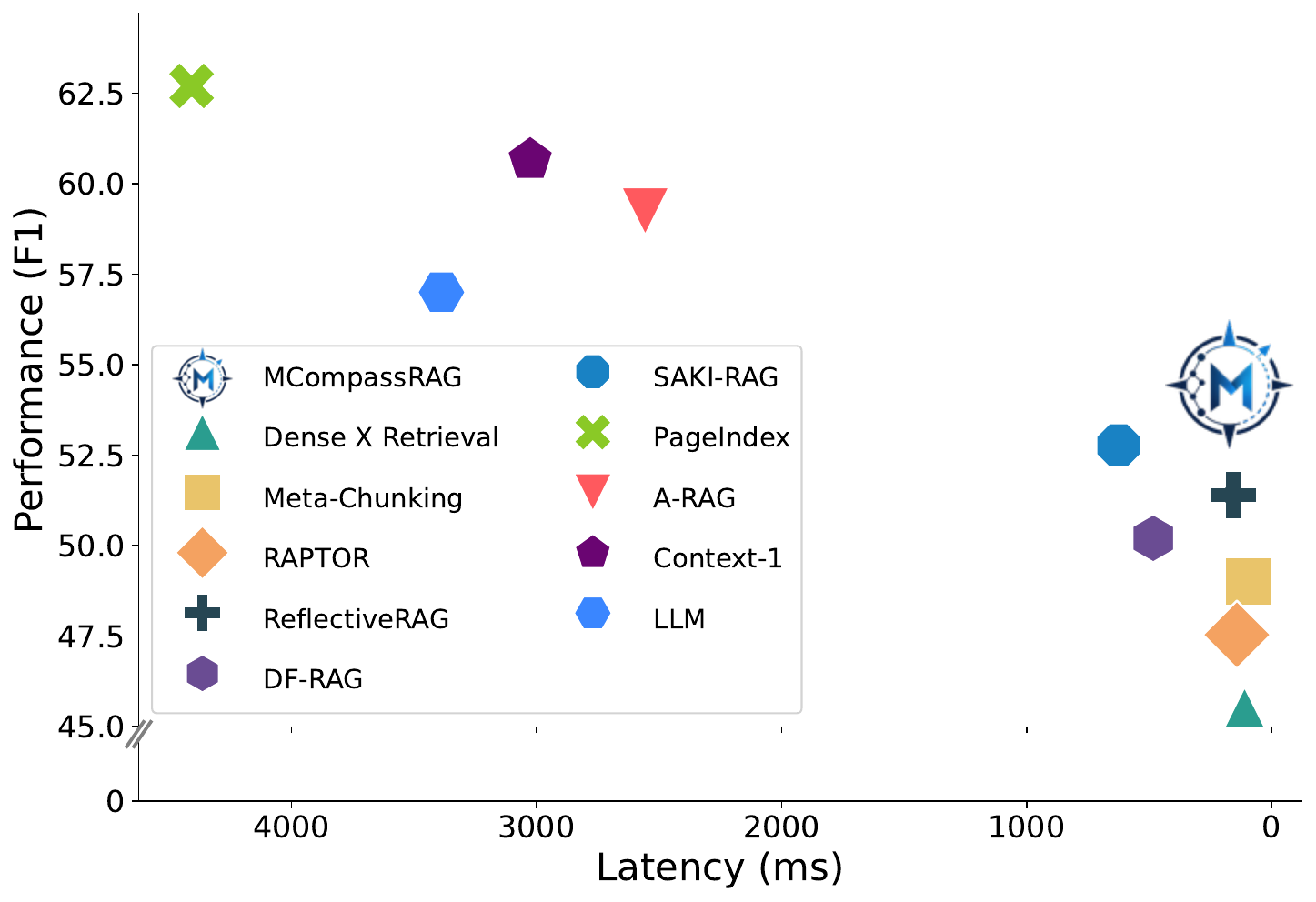}
        \caption{}
        \label{fig:accuracy-latency}
    \end{subfigure}
    \caption{
    Overview of \name{}. \textbf{(a)} \name{} uses coarse chunks for efficiency and enriches them with topic vectors for topic-aware retrieval. At query time, relevant topic information guides retrieval over larger chunks. \textbf{(b)} \name{} improves the performance--latency trade-off over strong RAG baselines, with performance measured by average F1 on HotpotQA~\cite{yang-etal-2018-hotpotqa} and DRBench~\cite{abaskohi2026drbench}.
    }
    \label{fig:main-teaser}
    \vspace{-1em}
\end{figure*}

Prior work addresses chunk granularity by either making chunks smaller, more structured, or hierarchically organized. Proposition-level retrieval decomposes documents into atomic units~\cite{chen-etal-2024-dense}, LLM-guided segmentation improves chunk boundaries~\cite{zhao2024meta, zhao-etal-2025-moc}, and hierarchical methods such as RAPTOR retrieve across multiple abstraction levels~\cite{sarthi2024raptor}. While effective, these approaches often increase pre-processing cost, require additional indices, or introduce extra scoring and selection stages. LLM-based re-ranking and evidence selection can further improve quality~\cite{tao-etal-2025-saki}, but add latency at inference time, which is problematic for deep research agents that repeatedly retrieve evidence over large corpora~\cite{zheng-etal-2025-deepresearcher}.

In this work, we take a different approach: rather than making chunks increasingly fine-grained, adding hierarchical retrieval stages, or relying on expensive post-retrieval filtering, we make coarse-grained chunks more searchable. As shown in Figure~\ref{fig:teaser}, \textbf{\name{}} enriches each chunk with topic metadata that acts as a semantic compass for retrieval. Specifically, a topic modeling encoder maps documents and chunks into topic-aware vectors in the same semantic space as the retriever. These topic vectors expose the main semantic directions covered by each coarse chunk, allowing retrieval to look beyond a single noisy chunk embedding. At query time, \name{} derives a compact query-side topic representation from the metadata bank and uses it to score metadata-enriched chunks. \name{} is agnostic to the specific topic model, requiring only that topics be embedded in the retriever's semantic space. We train \name{} as an extreme multi-label classifier~\cite{prabhu2025mogic} using LLM-teacher distillation, where a lightweight student learns to identify multiple relevant chunks from metadata-enriched representations without LLM calls at inference time. This preserves the efficiency advantage of larger chunks while reducing the semantic noise that makes coarse-grained cosine retrieval unreliable. Across six complex retrieval benchmarks, \name{} \textbf{improves information efficiency by 8.24\%} on average over the strongest non-LLM baseline while running at over \textbf{5$\times$ lower latency} compared to strong LLM-based RAG baselines, reflecting the efficiency--quality trade-off illustrated in Figure~\ref{fig:accuracy-latency}.

Our \textbf{contributions} are threefold. \textbf{First}, we introduce \textbf{\name{}}, a metadata-guided retrieval framework that improves coarse-grained retrieval by using selected topic metadata to make large chunks more precisely searchable without increasing the retrieval search space. \textbf{Second}, we design a metadata selection and abstraction mechanism that first selects the topical metadata most relevant to the query from a corpus-level metadata bank, then summarizes these signals into a compact query-topic vector used for chunk scoring. This makes the query representation topic-aware before matching it against coarse-grained chunks. \textbf{Third}, we distill an LLM teacher into a lightweight student retriever trained with an extreme multi-label objective, enabling efficient topic-aware evidence selection without inference-time LLM calls while preserving most teacher-guided retrieval quality.

%% file: sections/3_relatedwork.tex
\section{Related Work}

\begin{figure*}[h]
    \centering
    \includegraphics[width=\textwidth]{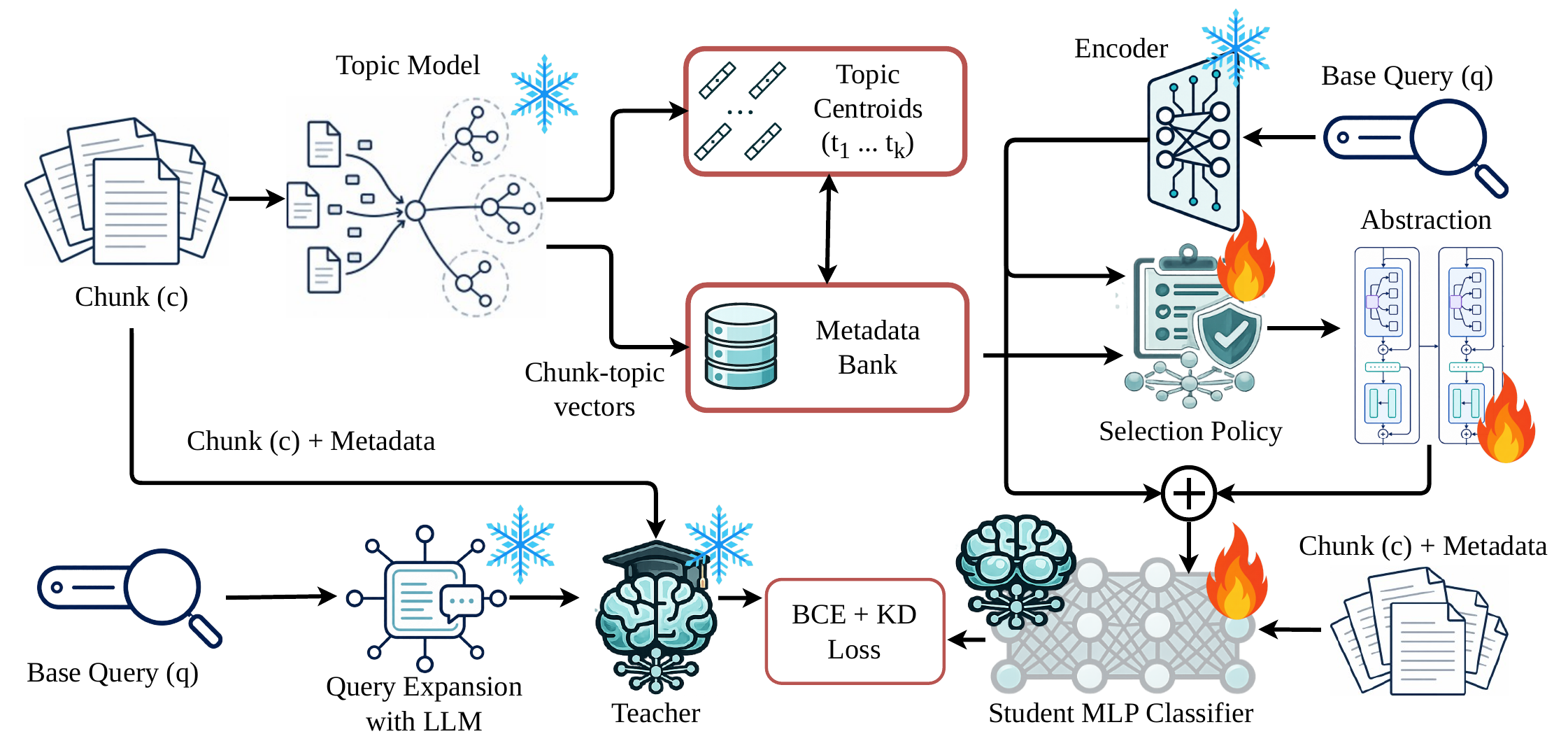}
    \caption{
    Overview of \name{}. During training, an LLM teacher provides relevance supervision, with query expansion used only as an additional teacher-side metadata signal. The metadata bank is built from chunks, enriched with document-topic vectors and topic centroid embeddings. At inference time, \name{} selects and abstracts query-relevant topic metadata, then scores query--chunk pairs with a lightweight student retriever. Icons indicate trainability: \faFire{} denotes trained components and \faSnowflake{} denotes frozen components.
    }
    \label{fig:architecture}
    \vspace{-1em}
\end{figure*}

\paragraph{Retrieval Granularity and Structured Retrieval in RAG.}
RAG grounds language model generation in external evidence retrieved before generation~\cite{lewis2020retrieval, karpukhin-etal-2020-dense, izacard-grave-2021-leveraging}. A key design choice is retrieval granularity: fine-grained units improve evidence precision but enlarge the search space and may lose context, while coarse-grained units preserve context and reduce candidates but make dense similarity noisier due to mixed topics and irrelevant content. Prior work addresses this trade-off through alternative retrieval units or index structures, including proposition-level retrieval~\cite{chen-etal-2024-dense}, LLM-guided and adaptive chunking~\cite{zhao2024meta, zhao-etal-2025-moc}, query-adaptive granularity selection~\cite{zhang2026smartchunk}, and hierarchical retrieval across abstraction levels~\cite{sarthi2024raptor}. Other systems enrich retrieved evidence to reduce context fragmentation~\cite{tao-etal-2025-saki} or promote diversity and coverage during selection~\cite{khan-etal-2026-df}. While effective, these methods often require finer-grained indexing, adaptive selection, hierarchical structures, extra scoring stages, or LLM-based filtering. In contrast, \name{} preserves the efficiency of coarse-grained retrieval while making larger chunks more searchable with topic-level metadata.

\paragraph{Semantic Guidance and Efficient Retrieval. }
A complementary line of work improves RAG by modifying the query or retrieval process rather than the chunking strategy itself. Query augmentation methods such as HyDE~\cite{gao-etal-2023-precise}, query expansion~\cite{wang-etal-2023-query2doc, zhou-etal-2024-hyqe}, and decomposition-based retrieval~\cite{trivedi-etal-2023-interleaving, zheng2024take} aim to better align the query with relevant evidence by generating hypothetical answers, adding related terms, or breaking complex questions into simpler retrieval steps. Adaptive and iterative retrieval methods further refine the evidence set through repeated retrieval, reranking, or sufficiency checking~\cite{verma-etal-2026-reflectiverag}. These methods are effective when the query underspecifies the needed evidence, but they often introduce extra inference-time computation. Separately, generation-side efficiency methods compress or reorganize retrieved context after retrieval to reduce decoding cost~\cite{lin2025refrag, louis2026oscar}. \name{} is orthogonal to these directions: rather than generating additional query text, repeatedly retrieving, or compressing context after retrieval, it uses corpus-derived topic metadata as a compact semantic guide before retrieval. This guides retrieval toward query-relevant topics without inference-time LLM calls, and remains compatible with query expansion, iterative retrieval, reranking, and context compression.

%% file: sections/4_method.tex
\section{\name{}}

\name{} is a metadata-guided retrieval framework that makes coarse-grained chunks more searchable without increasing the retrieval search space. Given chunks $\mathcal{C}=\{c_1,\ldots,c_N\}$ and a query $q$, the goal is to retrieve the top-$k$ chunks that provide useful evidence for answering the query. Instead of relying only on cosine similarity between query and chunk embeddings, \name{} augments both queries and chunks with topic-level metadata, allowing the retriever to better identify which semantic directions within a large chunk are relevant.

Figure~\ref{fig:architecture} illustrates the full pipeline. First, each chunk is processed by a topic model to obtain a chunk-topic distribution, while topic centroids provide embedding-space representations of the topics. The chunk-topic distributions are cached in a corpus-level metadata bank and later used as query-side guidance. At inference time, the base query is encoded by the student encoder, and a selection policy compares the query embedding with metadata entries from the bank to select the most relevant topic distributions. An abstraction module then summarizes the selected metadata distributions into a refined query-topic distribution, reducing noise and bias from any single selected entry. This refined distribution is converted into a compact query-side topic vector and concatenated with the query embedding to form the metadata-enriched query representation. The student MLP classifier then scores this representation against each metadata-enriched chunk representation and returns the top-$k$ chunks. During training, an LLM teacher provides relevance supervision using expanded queries, while the student receives only the base query and learns through BCE and knowledge-distillation losses. Thus, query expansion and LLM teacher scoring are used only for training; inference requires only metadata selection, abstraction, and student scoring. The framework can use any topic model whose topics are represented in the retriever embedding space and that provides chunk-level topic distributions. In our implementation, we use CEMTM~\cite{abaskohi-etal-2025-cemtm}, an LLM-distilled topic model that also leverages attention signals to produce document-topic distributions.

\subsection{Topic Metadata and Metadata Bank}
\label{sec:metadata_bank}

Let $\{\mathbf{t}_k\}_{k=1}^{K}$ denote the topic centroids, where each $\mathbf{t}_k \in \mathbb{R}^{d}$ lies in the retriever embedding space and serves as the vector representation, or prototype, of topic $k$. Each chunk $c$ is associated with a topic distribution $\boldsymbol{\theta}_c \in \mathbb{R}^{K}$, where $\theta_{c,r}$ measures the strength of topic $r$ in chunk $c$. Since chunks are longer and more informative than queries, their topic distributions can be computed reliably and cached offline. \name{} stores these chunk-level topic distributions in a metadata bank:

\begin{equation}
    \mathcal{M} = \{\boldsymbol{\theta}_{c_1}, \ldots, \boldsymbol{\theta}_{c_N}\}.
\end{equation}
The \textbf{metadata bank} represents the topical structure of the corpus and serves as the source of \textbf{query-side guidance at inference time}. Intuitively, it provides a corpus-level map of the semantic regions that queries may need to search, without relying only on the sparse signal in the query itself. Given a new query, \name{} does not directly rely on the query's own topic distribution, which may be unreliable due to its short length. Instead, it selects relevant topic distributions from $\mathcal{M}$ and abstracts them into a compact query-side topic representation. This abstraction step reduces bias toward any single selected chunk and produces a smoother topical signal, as described in Section~\ref{sec:selection_representation}.

\subsection{Metadata Selection and Representation}
\label{sec:selection_representation}

At inference time, \name{} selects topic metadata from the bank that is relevant to the input query. The query is first encoded by the student encoder, $f_{\psi}$:
\begin{equation}
    \mathbf{e}_q = f_{\psi}(q) \in \mathbb{R}^{d}.
\end{equation}
We implement the \textbf{selection policy} as a lightweight scoring module over the concatenation of the query embedding and each metadata-entry embedding. Each metadata entry $\boldsymbol{\theta}_{c_i}$ is first converted into an embedding-space summary:
\begin{equation}
    \mathbf{m}_i = \sum_{k=1}^{K} \theta_{c_i,k}\mathbf{t}_k .
\end{equation}
The selector then assigns an unnormalized compatibility score between the query embedding and each metadata-entry summary:
\begin{equation}
    a_i = \mathbf{w}_s^\top [\mathbf{e}_q ; \mathbf{m}_i] + b_s ,
\end{equation}
where $[\cdot;\cdot]$ denotes concatenation. The scores are converted into a probability distribution over metadata entries using a softmax operation:
\begin{equation}
    s_i =
    \frac{\exp(a_i)}
    {\sum_{j=1}^{N} \exp(a_j)} .
\end{equation}
The top-$L$ metadata entries according to $s_i$ are selected and passed to the \textbf{abstraction module}.
\begin{equation}
    \mathbf{H}^{(0)}
    =
    [\boldsymbol{\theta}_{c_{j_1}};\ldots;\boldsymbol{\theta}_{c_{j_L}}]
    \in \mathbb{R}^{L \times K}.
\end{equation}
After a two-layer Transformer encoder~\cite{NIPS2017_3f5ee243}, the outputs are mean-pooled to form a refined query topic distribution:
\begin{equation}
    \hat{\boldsymbol{\theta}}_q =
    \frac{1}{L}\sum_{\ell=1}^{L} \mathbf{H}^{(2)}_{\ell}.
\end{equation}
This abstraction step combines complementary topic signals and suppresses redundant or noisy metadata entries and constructs topic-enriched representations for both chunks and queries. For a chunk $c$, we select the top-$M$ topics from its topic distribution (here, $L$ is the number of selected metadata entries, while $M$ is the number of selected topics):
\begin{equation}
    \mathcal{T}_c = \operatorname{top\text{-}M}(\boldsymbol{\theta}_c),
\end{equation}
and aggregate their topic centroids:
\begin{equation}
    \mathbf{g}_c =
    \sum_{k \in \mathcal{T}_c}
    \theta_{c,k}\mathbf{t}_k .
\end{equation}
The final chunk representation is $\mathbf{r}_c = [\mathbf{e}_c ; \mathbf{g}_c]$, where $\mathbf{e}_c=f_{\psi}(c)$ is the chunk embedding produced by the student encoder. Similarly, the refined query topic distribution $\hat{\boldsymbol{\theta}}_q$ is used to build a query-side topic summary with the top-$M$ topics, yielding $\mathbf{r}_q = [\mathbf{e}_q ; \mathbf{g}_q]$.

The student retriever scores each query--chunk pair with a three-layer MLP classifier:
\begin{equation}
    z(q,c) =
    \operatorname{MLP}_{\phi}([\mathbf{r}_q ; \mathbf{r}_c]),
\end{equation}
where $z(q,c)$ is the predicted relevance logit. This formulation casts retrieval as an extreme multi-label classification problem: each chunk is a candidate label, and each query may correspond to multiple relevant chunks.

\subsection{Training with LLM-Teacher Distillation}
\label{sec:training}

\noindent
\textbf{Training data construction. }
We synthesize training data from the training split of each benchmark. For each dataset, we sample 2{,}000 chunks and use \textsc{GPT-4o}~\cite{openai2024gpt4o} to generate 10 natural queries per chunk, resulting in 20{,}000 query--chunk pairs before negative sampling. For each sampled chunk $c_i$, \textsc{GPT-4o} receives the target chunk together with its preceding and following chunks. It first generates a base query $q_i$ whose answer requires evidence from $c_i$. It then generates an expanded query $\tilde{q}_i$ by adding only background information from the two of the neighboring chunks, without revealing the answer or including answer-specific hints. We use Prompt~\hyperref[app:training_prompts]{\ref*{app:training_prompts}.1} for the query expansion.

\noindent
\textbf{Training procedure and objective. }
For relevance supervision, the source chunk is treated as a positive candidate, while negatives are sampled from non-matching chunks. We include both random negatives and hard negatives, where hard negatives are retrieved using \texttt{Qwen3-Embedding-4B}~\cite{zhang2025qwen3embeddingadvancingtext} as high-similarity chunks that the LLM teacher judges as not useful for answering the query. GPT-4o is then used as an LLM teacher: given the expanded query $\tilde{q}_i$ and a candidate chunk, it predicts whether the chunk provides direct or supporting evidence for answering the query (see Prompt~\hyperref[app:training_prompts]{\ref*{app:training_prompts}.2}). The resulting hard label $y \in \{0,1\}$ and teacher score/logit $z^{\mathrm{T}}$ are used as supervision for the student relevance classifier.

The teacher scores each query--chunk pair using the expanded query $\tilde{q}_i$, whereas the student receives only the base query $q_i$. This information asymmetry encourages the student to recover useful missing context through metadata selection and abstraction. The training objective combines hard-label binary cross-entropy with soft teacher distillation:
\begin{equation}
    \mathcal{L}
    =
    (1-\alpha)\mathcal{L}_{\mathrm{BCE}}
    +
    \alpha \mathcal{L}_{\mathrm{KD}},
\end{equation}
where $\alpha$ balances hard-label learning and soft distillation. The binary cross-entropy loss is
\begin{equation}
    \mathcal{L}_{\mathrm{BCE}}
    =
    - y \log \sigma(z)
    - (1-y)\log(1-\sigma(z)),
\end{equation}
where $z$ is the student relevance logit and $\sigma$ is the sigmoid function. The distillation term matches the teacher and student soft scores:
\begin{equation}
    \mathcal{L}_{\mathrm{KD}}
    =
    \mathrm{KL}
    \left(
    \sigma(z^{\mathrm{T}}/\tau)
    \;\|\;
    \sigma(z/\tau)
    \right),
\end{equation}
where $z^{\mathrm{T}}$ is the teacher score/logit and $\tau$ is the temperature. The student encoder, topic centroids, and cached chunk topic distributions are kept fixed. We train only the metadata selector, abstraction module, and MLP relevance classifier.

\begin{algorithm}[t]
\caption{\name{} Inference}
\label{alg:inference}
\begin{algorithmic}[1]
\REQUIRE Query $q$, precomputed chunk representations $\{\mathbf{r}_{c_j}\}_{j=1}^{N}$, metadata bank $\mathcal{M}$, topic centroids $\{\mathbf{t}_r\}_{r=1}^{K}$, selected metadata count $L$, top topics $M$, retrieved chunks $k$
\ENSURE Retrieved chunk set $\mathcal{C}_k$

\STATE $\mathbf{e}_q \leftarrow f_{\psi}(q)$

\STATE \textbf{// Metadata selection}
\FOR{each metadata entry $\boldsymbol{\theta}_{c_i} \in \mathcal{M}$}
    \STATE $\mathbf{m}_i \leftarrow \sum_{r=1}^{K} \theta_{c_i,r}\mathbf{t}_r$
    \STATE $a_i \leftarrow \mathbf{w}_s^\top [\mathbf{e}_q ; \mathbf{m}_i] + b_s$
\ENDFOR
\STATE $s_i \leftarrow \frac{\exp(a_i)}{\sum_{j=1}^{|\mathcal{M}|}\exp(a_j)}$
\STATE $\mathcal{S} \leftarrow \operatorname{top\text{-}L}(\{s_i\})$

\STATE \textbf{// Metadata abstraction}
\STATE $\mathbf{H}^{(0)} \leftarrow [\boldsymbol{\theta}_{c_j}]_{j \in \mathcal{S}}$
\STATE $\hat{\boldsymbol{\theta}}_q \leftarrow \operatorname{MeanPool}(\operatorname{TransformerEnc}(\mathbf{H}^{(0)}))$
\STATE $\mathcal{T}_q \leftarrow \operatorname{top\text{-}M}(\hat{\boldsymbol{\theta}}_q)$
\STATE $\mathbf{g}_q \leftarrow \sum_{r \in \mathcal{T}_q} \hat{\theta}_{q,r}\mathbf{t}_r$
\STATE $\mathbf{r}_q \leftarrow [\mathbf{e}_q;\mathbf{g}_q]$

\STATE \textbf{// Retrieval}
\FOR{each precomputed $\mathbf{r}_{c_j}$}
    \STATE $z_j \leftarrow \operatorname{MLP}_{\phi}([\mathbf{r}_q ; \mathbf{r}_{c_j}])$
\ENDFOR
\STATE $\mathcal{C}_k \leftarrow \operatorname{top\text{-}k}_{c_j \in \mathcal{C}}(\{z_j\})$

\RETURN $\mathcal{C}_k$
\end{algorithmic}
\end{algorithm}

\subsection{Inference}
\label{sec:inference}

At inference time, \name{} retrieves evidence without LLM calls. All chunk embeddings, topic distributions, and topic-enriched chunk representations are precomputed offline as indices for retrieval. For a given query, \name{} computes the query embedding, selects and abstracts relevant metadata from the bank, scores all cached chunks with the MLP classifier, and returns the top-$k$ results. Algorithm~\ref{alg:inference} summarizes this procedure. Since topic extraction and chunk encoding are offline, online inference only requires lightweight metadata selection, abstraction, and scoring.

%% file: sections/5_experiments.tex
\section{Experiments and Results}

\subsection{Experimental Setup}
\label{sec:experimental_setup}

\noindent
\textbf{Models and implementation. }
We use \textsc{Qwen3-Embedding-4B}~\citep{zhang2025qwen3embeddingadvancingtext} as the student encoder for query and chunk representations, and \textsc{Qwen3-32B}~\citep{qwen3technicalreport} as both the LLM teacher for relevance supervision and the final answer generator. For baselines requiring LLM-based generation, planning, or selection, we use the same LLM scale for fair comparison. When a baseline requires reranking, we use \textsc{Qwen3-Reranker-4B}~\citep{zhang2025qwen3embeddingadvancingtext}. Closed-source API-based components are accessed through OpenRouter\footnote{\url{https://openrouter.ai/}}. All experiments are run with access to 8 NVIDIA A100 80GB GPUs.

\noindent
\textbf{Topic metadata. }
We use CEMTM~\cite{abaskohi-etal-2025-cemtm} with \textsc{Qwen3-Embedding-4B} as the topic modeling backbone. CEMTM is trained on WikiWeb2M~\cite{burns2023wiki} with $K=100$ topics. See Appendix~\ref{app:topic_granularity} for the topic granularity analysis. We use only the CEMTM encoder to obtain chunk-level document-topic vectors and topic centroid embeddings. Since the LLM teacher also requires topic-aware representations, we additionally use a \textsc{Qwen3-32B}-based CEMTM variant for teacher-side topic modeling. We ablate the in-domain topic modeling in Appendix~\ref{app:in_domain_topic_model}.

\noindent
\textbf{Benchmarks. }
We evaluate on seven benchmarks: SCI-DOCS~\cite{cohan-etal-2020-specter}, LegalBench-RAG~\cite{pipitone2024legalbenchragbenchmarkretrievalaugmentedgeneration}, Dragonball~\cite{zhu-etal-2025-rageval}, HotpotQA~\cite{yang-etal-2018-hotpotqa}, SQuAD~\cite{rajpurkar-etal-2016-squad}, DRBench~\cite{abaskohi2026drbench}, and LongBenchV2~\cite{bai-etal-2025-longbench}. For retrieval evaluation, we use SCI-DOCS, LegalBench-RAG, Dragonball, HotpotQA, SQuAD, and DRBench, which provide evidence annotations or links convertible to chunk-level labels. We use LongBenchV2 only for downstream evaluation, as it lacks chunk-level evidence labels. See Appendix~\ref{app:benchmark_baseline_details} for more details.

\begin{table*}[h]
    \centering
    \setlength{\tabcolsep}{3.2pt}
    \renewcommand{\arraystretch}{1.1}
    \resizebox{\textwidth}{!}{%
    \begin{tabular}{lccccccccc}
    \toprule
    \multirow{2}{*}{\textbf{Method}}
      & \multicolumn{3}{c}{\textbf{Dragonball}}
      & \multicolumn{3}{c}{\textbf{HotpotQA}}
      & \multicolumn{3}{c}{\textbf{SQuAD}} \\
    \cmidrule(lr){2-4}\cmidrule(lr){5-7}\cmidrule(lr){8-10}
      & \textit{IE$\uparrow$} & \textit{Prec.$\uparrow$} & \textit{Rec.$\uparrow$}
      & \textit{IE$\uparrow$} & \textit{Prec.$\uparrow$} & \textit{Rec.$\uparrow$}
      & \textit{IE$\uparrow$} & \textit{Prec.$\uparrow$} & \textit{Rec.$\uparrow$} \\
    \midrule
    RAPTOR
      & 30.13{\scriptsize\textcolor{red!70!black}{$\pm$.41}} & 39.40{\scriptsize\textcolor{red!70!black}{$\pm$.52}} & 10.53{\scriptsize\textcolor{red!70!black}{$\pm$.29}}
      & 45.43{\scriptsize\textcolor{red!70!black}{$\pm$.63}} & 59.63{\scriptsize\textcolor{red!70!black}{$\pm$.58}} & 13.70{\scriptsize\textcolor{red!70!black}{$\pm$.34}}
      & 60.70{\scriptsize\textcolor{red!70!black}{$\pm$.71}} & 32.77{\scriptsize\textcolor{red!70!black}{$\pm$.44}} & 21.13{\scriptsize\textcolor{red!70!black}{$\pm$.39}} \\
    Meta-Chunking-MSP
      & 31.40{\scriptsize\textcolor{red!70!black}{$\pm$.38}} & 40.20{\scriptsize\textcolor{red!70!black}{$\pm$.47}} & 11.63{\scriptsize\textcolor{red!70!black}{$\pm$.31}}
      & 55.70{\scriptsize\textcolor{red!70!black}{$\pm$.69}} & 64.30{\scriptsize\textcolor{red!70!black}{$\pm$.62}} & 17.97{\scriptsize\textcolor{red!70!black}{$\pm$.42}}
      & 80.60{\scriptsize\textcolor{red!70!black}{$\pm$.58}} & 41.97{\scriptsize\textcolor{red!70!black}{$\pm$.53}} & 34.40{\scriptsize\textcolor{red!70!black}{$\pm$.49}} \\
    Meta-Chunking-PPL
      & 40.87{\scriptsize\textcolor{red!70!black}{$\pm$.45}} & 42.80{\scriptsize\textcolor{red!70!black}{$\pm$.50}} & 15.73{\scriptsize\textcolor{red!70!black}{$\pm$.36}}
      & 66.77{\scriptsize\textcolor{red!70!black}{$\pm$.73}} & 65.23{\scriptsize\textcolor{red!70!black}{$\pm$.64}} & 21.40{\scriptsize\textcolor{red!70!black}{$\pm$.47}}
      & 78.80{\scriptsize\textcolor{red!70!black}{$\pm$.62}} & 41.37{\scriptsize\textcolor{red!70!black}{$\pm$.55}} & 33.70{\scriptsize\textcolor{red!70!black}{$\pm$.51}} \\
    DenseXRetrieval
      & 2.27{\scriptsize\textcolor{red!70!black}{$\pm$.12}} & 4.40{\scriptsize\textcolor{red!70!black}{$\pm$.18}} & 0.09{\scriptsize\textcolor{red!70!black}{$\pm$.03}}
      & 35.60{\scriptsize\textcolor{red!70!black}{$\pm$.56}} & 43.17{\scriptsize\textcolor{red!70!black}{$\pm$.49}} & 7.03{\scriptsize\textcolor{red!70!black}{$\pm$.21}}
      & 61.53{\scriptsize\textcolor{red!70!black}{$\pm$.68}} & 31.17{\scriptsize\textcolor{red!70!black}{$\pm$.46}} & 19.83{\scriptsize\textcolor{red!70!black}{$\pm$.37}} \\
    SAKI-RAG
      & 32.90{\scriptsize\textcolor{red!70!black}{$\pm$.42}} & 71.37{\scriptsize\textcolor{red!70!black}{$\pm$.66}} & 25.40{\scriptsize\textcolor{red!70!black}{$\pm$.45}}
      & 58.73{\scriptsize\textcolor{red!70!black}{$\pm$.70}} & 55.60{\scriptsize\textcolor{red!70!black}{$\pm$.59}} & 30.03{\scriptsize\textcolor{red!70!black}{$\pm$.52}}
      & 87.17{\scriptsize\textcolor{red!70!black}{$\pm$.51}} & 88.80{\scriptsize\textcolor{red!70!black}{$\pm$.43}} & 78.93{\scriptsize\textcolor{red!70!black}{$\pm$.57}} \\
    \hline
    LLM
      & 34.73{\scriptsize\textcolor{red!70!black}{$\pm$.39}} & 76.53{\scriptsize\textcolor{red!70!black}{$\pm$.61}} & 27.30{\scriptsize\textcolor{red!70!black}{$\pm$.43}}
      & 62.63{\scriptsize\textcolor{red!70!black}{$\pm$.67}} & 55.83{\scriptsize\textcolor{red!70!black}{$\pm$.55}} & 33.50{\scriptsize\textcolor{red!70!black}{$\pm$.49}}
      & 89.93{\scriptsize\textcolor{red!70!black}{$\pm$.46}} & 91.63{\scriptsize\textcolor{red!70!black}{$\pm$.40}} & 82.77{\scriptsize\textcolor{red!70!black}{$\pm$.52}} \\
    LLM + 10 Topics
      & \textbf{40.83}{\scriptsize\textcolor{red!70!black}{$\pm$.34}} & \textbf{87.43}{\scriptsize\textcolor{red!70!black}{$\pm$.49}} & \textbf{34.17}{\scriptsize\textcolor{red!70!black}{$\pm$.38}}
      & \textbf{72.90}{\scriptsize\textcolor{red!70!black}{$\pm$.58}} & \textbf{59.33}{\scriptsize\textcolor{red!70!black}{$\pm$.51}} & \textbf{42.70}{\scriptsize\textcolor{red!70!black}{$\pm$.44}}
      & \textbf{94.10}{\scriptsize\textcolor{red!70!black}{$\pm$.33}} & \textbf{95.83}{\scriptsize\textcolor{red!70!black}{$\pm$.29}} & \textbf{89.50}{\scriptsize\textcolor{red!70!black}{$\pm$.36}} \\
    \hline
    \rowcolor{oursblue}
    MCompassRAG + 10 Topics
      & \underline{38.97}{\scriptsize\textcolor{red!70!black}{$\pm$.36}} & \underline{82.80}{\scriptsize\textcolor{red!70!black}{$\pm$.52}} & \underline{32.40}{\scriptsize\textcolor{red!70!black}{$\pm$.40}}
      & \underline{70.17}{\scriptsize\textcolor{red!70!black}{$\pm$.61}} & \underline{56.40}{\scriptsize\textcolor{red!70!black}{$\pm$.48}} & \underline{40.63}{\scriptsize\textcolor{red!70!black}{$\pm$.46}}
      & \underline{93.80}{\scriptsize\textcolor{red!70!black}{$\pm$.35}} & \underline{95.37}{\scriptsize\textcolor{red!70!black}{$\pm$.31}} & \underline{88.90}{\scriptsize\textcolor{red!70!black}{$\pm$.38}} \\
    \bottomrule
    \end{tabular}
    }% end resizebox

    \vspace{2pt}

    \resizebox{\textwidth}{!}{%
    \begin{tabular}{lccccccccc}
    \toprule
    \multirow{2}{*}{\textbf{Method}}
      & \multicolumn{3}{c}{\textbf{DRBench}}
      & \multicolumn{3}{c}{\textbf{LegalBench-RAG}}
      & \multicolumn{3}{c}{\textbf{SCI-DOCS}} \\
    \cmidrule(lr){2-4}\cmidrule(lr){5-7}\cmidrule(lr){8-10}
      & \textit{IE$\uparrow$} & \textit{Prec.$\uparrow$} & \textit{Rec.$\uparrow$}
      & \textit{IE$\uparrow$} & \textit{Prec.$\uparrow$} & \textit{Rec.$\uparrow$}
      & \textit{IE$\uparrow$} & \textit{Prec.$\uparrow$} & \textit{Rec.$\uparrow$} \\
    \midrule
    RAPTOR
      & 24.13{\scriptsize\textcolor{red!70!black}{$\pm$.37}} & 32.77{\scriptsize\textcolor{red!70!black}{$\pm$.44}} &  8.20{\scriptsize\textcolor{red!70!black}{$\pm$.25}}
      & 24.27{\scriptsize\textcolor{red!70!black}{$\pm$.35}} & 32.23{\scriptsize\textcolor{red!70!black}{$\pm$.42}} &  8.20{\scriptsize\textcolor{red!70!black}{$\pm$.24}}
      & 88.63{\scriptsize\textcolor{red!70!black}{$\pm$.54}} & 82.77{\scriptsize\textcolor{red!70!black}{$\pm$.50}} & 80.37{\scriptsize\textcolor{red!70!black}{$\pm$.55}} \\
    Meta-Chunking-MSP
      & 30.60{\scriptsize\textcolor{red!70!black}{$\pm$.42}} & 36.13{\scriptsize\textcolor{red!70!black}{$\pm$.47}} & 12.30{\scriptsize\textcolor{red!70!black}{$\pm$.31}}
      & 28.30{\scriptsize\textcolor{red!70!black}{$\pm$.39}} & 36.10{\scriptsize\textcolor{red!70!black}{$\pm$.45}} & 11.07{\scriptsize\textcolor{red!70!black}{$\pm$.29}}
      & 90.47{\scriptsize\textcolor{red!70!black}{$\pm$.49}} & 83.53{\scriptsize\textcolor{red!70!black}{$\pm$.48}} & 82.10{\scriptsize\textcolor{red!70!black}{$\pm$.52}} \\
    Meta-Chunking-PPL
      & 36.30{\scriptsize\textcolor{red!70!black}{$\pm$.49}} & 37.57{\scriptsize\textcolor{red!70!black}{$\pm$.51}} & 16.17{\scriptsize\textcolor{red!70!black}{$\pm$.34}}
      & 32.70{\scriptsize\textcolor{red!70!black}{$\pm$.43}} & 37.53{\scriptsize\textcolor{red!70!black}{$\pm$.48}} & 13.57{\scriptsize\textcolor{red!70!black}{$\pm$.32}}
      & 21.07{\scriptsize\textcolor{red!70!black}{$\pm$.36}} & 17.60{\scriptsize\textcolor{red!70!black}{$\pm$.31}} &  3.57{\scriptsize\textcolor{red!70!black}{$\pm$.15}} \\
    DenseXRetrieval
      & 18.40{\scriptsize\textcolor{red!70!black}{$\pm$.31}} & 25.37{\scriptsize\textcolor{red!70!black}{$\pm$.38}} &  5.43{\scriptsize\textcolor{red!70!black}{$\pm$.19}}
      & 19.53{\scriptsize\textcolor{red!70!black}{$\pm$.33}} & 24.93{\scriptsize\textcolor{red!70!black}{$\pm$.36}} &  5.13{\scriptsize\textcolor{red!70!black}{$\pm$.18}}
      & 86.00{\scriptsize\textcolor{red!70!black}{$\pm$.57}} & 79.33{\scriptsize\textcolor{red!70!black}{$\pm$.53}} & 74.67{\scriptsize\textcolor{red!70!black}{$\pm$.60}} \\
    SAKI-RAG
      & 37.47{\scriptsize\textcolor{red!70!black}{$\pm$.46}} & 62.30{\scriptsize\textcolor{red!70!black}{$\pm$.61}} & 28.23{\scriptsize\textcolor{red!70!black}{$\pm$.43}}
      & 31.23{\scriptsize\textcolor{red!70!black}{$\pm$.41}} & 46.30{\scriptsize\textcolor{red!70!black}{$\pm$.52}} & 19.27{\scriptsize\textcolor{red!70!black}{$\pm$.36}}
      & 86.53{\scriptsize\textcolor{red!70!black}{$\pm$.50}} & 92.27{\scriptsize\textcolor{red!70!black}{$\pm$.43}} & 84.30{\scriptsize\textcolor{red!70!black}{$\pm$.51}} \\
    \hline
    LLM
      & 41.53{\scriptsize\textcolor{red!70!black}{$\pm$.44}} & 68.43{\scriptsize\textcolor{red!70!black}{$\pm$.57}} & 32.27{\scriptsize\textcolor{red!70!black}{$\pm$.41}}
      & 33.93{\scriptsize\textcolor{red!70!black}{$\pm$.39}} & 50.40{\scriptsize\textcolor{red!70!black}{$\pm$.49}} & 22.13{\scriptsize\textcolor{red!70!black}{$\pm$.35}}
      & 89.37{\scriptsize\textcolor{red!70!black}{$\pm$.45}} & 95.10{\scriptsize\textcolor{red!70!black}{$\pm$.34}} & 87.47{\scriptsize\textcolor{red!70!black}{$\pm$.43}} \\
    LLM + 10 Topics
      & \textbf{50.27}{\scriptsize\textcolor{red!70!black}{$\pm$.39}} & \textbf{83.17}{\scriptsize\textcolor{red!70!black}{$\pm$.46}} & \textbf{43.30}{\scriptsize\textcolor{red!70!black}{$\pm$.37}}
      & \textbf{40.10}{\scriptsize\textcolor{red!70!black}{$\pm$.34}} & \textbf{59.47}{\scriptsize\textcolor{red!70!black}{$\pm$.43}} & \textbf{29.70}{\scriptsize\textcolor{red!70!black}{$\pm$.31}}
      & \textbf{94.67}{\scriptsize\textcolor{red!70!black}{$\pm$.30}} & \textbf{99.50}{\scriptsize\textcolor{red!70!black}{$\pm$.12}} & \textbf{92.50}{\scriptsize\textcolor{red!70!black}{$\pm$.28}} \\
    \hline
    \rowcolor{oursblue}
    MCompassRAG + 10 Topics
      & \underline{47.97}{\scriptsize\textcolor{red!70!black}{$\pm$.41}} & \underline{78.57}{\scriptsize\textcolor{red!70!black}{$\pm$.49}} & \underline{41.20}{\scriptsize\textcolor{red!70!black}{$\pm$.39}}
      & \underline{38.40}{\scriptsize\textcolor{red!70!black}{$\pm$.36}} & \underline{55.10}{\scriptsize\textcolor{red!70!black}{$\pm$.45}} & \underline{27.90}{\scriptsize\textcolor{red!70!black}{$\pm$.33}}
      & \underline{94.13}{\scriptsize\textcolor{red!70!black}{$\pm$.32}} & \underline{99.03}{\scriptsize\textcolor{red!70!black}{$\pm$.15}} & \underline{92.10}{\scriptsize\textcolor{red!70!black}{$\pm$.29}} \\
    \bottomrule
    \end{tabular}
    }% end resizebox

    \caption{Retrieval performance across six benchmarks, averaged over three runs. \textcolor{red!70!black}{$\pm$ values} denote standard deviation. \textbf{Bold} = best; \underline{underline} = second-best; shaded rows indicate \name{}. LLM-based rows are inference-time oracle upper bounds. Detailed $k{=}1,3,5$ results are in Appendix~\ref{app:detailed_results}.}
    \label{tab:main_results}
    \vspace{-1em}
\end{table*}

\begin{table*}[t]
    \centering
    \resizebox{\textwidth}{!}{%
    \begin{tabular}{l
      S[table-format=2.1]
      S[table-format=2.1] S[table-format=2.1]
      S[table-format=2.1]
      S[table-format=1.3] S[table-format=1.3] S[table-format=1.3]
      S[table-format=5.0]
      S[table-format=4.0]
      c
      c}
    \toprule
    & \multicolumn{7}{c}{\textbf{Final Performance}}
    & \multicolumn{2}{c}{\textbf{}} \\
    \cmidrule(lr){2-8}
    & \multicolumn{1}{c}{\textbf{HotpotQA}}
    & \multicolumn{2}{c}{\textbf{LongBench v2}}
    & \multicolumn{1}{c}{\textbf{DRBench}}
    & \multicolumn{3}{c}{\textbf{Dragonball}}
    & \multicolumn{2}{c}{\textbf{Average Cost}} \\
    \cmidrule(lr){2-2} \cmidrule(lr){3-4} \cmidrule(lr){5-5} \cmidrule(lr){6-8} \cmidrule(lr){9-10}
    \textbf{Method}
      & {F1 $\uparrow$}
      & {F1 $\uparrow$} & {Acc $\uparrow$}
      & {F1 $\uparrow$}
      & {R-L $\uparrow$} & {MTR $\uparrow$} & {BRT $\uparrow$}
      & {Tok/Q $\downarrow$}
      & {Lat.\ (ms) $\downarrow$} \\
    \midrule

    \rowcolor{gray!20}
    \multicolumn{10}{c}{\textit{\textbf{Efficient RAG Methods}}} \\
    \addlinespace[2pt]
    Dense X Retrieval   & 60.9 & 26.4 & 28.6 & 46.8 & 0.248 & 0.269 & 0.548 & 2759  & 112  \\
    Meta-Chunking-PPL       & 64.5 & 29.7 & 31.8 & 50.7 & 0.272 & 0.292 & 0.571 & 2394  & 95   \\
    RAPTOR              & 63.1 & 28.3 & 30.4 & 49.1 & 0.264 & 0.285 & 0.563 & 3183  & 145  \\
    ReflectiveRAG       & 67.4 & 31.5 & 33.4 & 53.4 & 0.303 & 0.325 & 0.604 & 3527  & 161  \\
    DF-RAG              & 66.2 & 30.2 & 32.3 & 52.1 & 0.291 & 0.313 & 0.592 & 4843  & 484  \\
    SAKI-RAG            & 68.6 & 32.6 & 34.5 & 55.2 & 0.314 & 0.336 & 0.619 & 5584  & 925  \\
    REFRAG              & 73.6 & 37.5 & 39.4 & 60.4 & 0.354 & 0.371 & 0.650 & 7800  & 720  \\
    \midrule

    \rowcolor{gray!20}
    \multicolumn{10}{c}{\textit{\textbf{Long-Context Methods}}} \\
    \addlinespace[2pt]
    PageIndex           & 78.7 & 41.9 & 43.6 & 65.8 & 0.372 & 0.394 & 0.682 & 53883 & 4408 \\
    A-RAG               & 74.9 & 38.7 & 40.4 & 62.4 & 0.347 & 0.369 & 0.655 & 14625 & 2557 \\
    Chroma Context-1           & 76.1 & 40.1 & 41.8 & 64.1 & 0.359 & 0.382 & 0.669 & 20430 & 3026 \\
    LLM  & 72.9 & 36.9 & 38.8 & 59.3 & 0.352 & 0.362 & 0.642 & 41058 & 3388 \\
    \midrule

    \rowcolor{gray!20}
    \multicolumn{10}{c}{\textit{\textbf{Ours}}} \\
    \addlinespace[2pt]
    \rowcolor{oursblue}
    MCompassRAG         & 71.8 & 35.8 & 35.7 & 58.9 & 0.333 & 0.355 & 0.635 & 4126  & 174  \\
    \bottomrule
    \end{tabular}%
    }
    \caption{Downstream performance and efficiency across four benchmarks. We report task-specific generation metrics: Accuracy/F1 for QA-style datasets and ROUGE-L (R-L), METEOR (MTR), and BERTScore (BRT) for free-form generation. Tok/Q denotes the average retrieved tokens per query, and Lat.\ denotes end-to-end latency.}
    \label{tab:performance_results}
\end{table*}

\noindent
\textbf{Baselines. }
We compare against dense, structured, long-context, and LLM-based RAG baselines: DenseXRetrieval~\cite{chen-etal-2024-dense}, Meta-Chunking with PPL and MSP variants~\cite{zhao2024meta}, RAPTOR~\cite{sarthi2024raptor}, ReflectiveRAG~\cite{verma-etal-2026-reflectiverag}, DF-RAG~\cite{khan-etal-2026-df}, SAKI-RAG~\cite{tao-etal-2025-saki}, REFRAG~\cite{lin2025refrag}, PageIndex~\cite{zhang2025pageindex}, A-RAG~\cite{du2026aragscalingagenticretrievalaugmented}, Chroma Context-1~\cite{bashir2026context1}, and a long-context \textsc{Qwen3-32B} baseline. For retrieval evaluation, we include DenseXRetrieval, Meta-Chunking, RAPTOR, SAKI-RAG, and LLM retrievers, with both topic-free and topic-guided LLM variants. Other baselines are evaluated only downstream, as they mainly target generation, decoding, reranking, or context-use efficiency rather than standalone retrieval. Refer to Appendix~\ref{app:benchmark_baseline_details} for more details.

\noindent
\textbf{Training and evaluation.}
We train \name{} separately for each benchmark, using synthetic training data when retrieval labels are unavailable or insufficient; for DRBench and LongBenchV2, we train on EDR-200~\cite{prabhakar2025enterprisedeepresearchsteerable} and LongBenchV1~\cite{bai-etal-2024-longbench}, respectively. We train only the metadata selector, abstraction module, and MLP classifier, while keeping all encoders and cached topic representations fixed. Retrieval quality is measured by Recall, Precision, and Information Efficiency (IE), with $\mathrm{IE@k}=\mathrm{Precision@k}\times\mathrm{Recall@k}$, averaged over $k\in\{1,3,5\}$ and three runs. Downstream performance is evaluated with task-appropriate metrics: Accuracy, F1, ROUGE-L~\cite{lin-2004-rouge}, METEOR~\cite{banerjee-lavie-2005-meteor}, and BERTScore~\cite{Zhang2020BERTScore}. For fair comparison across chunk granularities, retrieved chunks are added in ranked order until a fixed token budget is reached (1K). Full training hyperparameters, inference, and evaluation settings are provided in Appendix~\ref{app:training_and_implementation}.

\subsection{Comparison with Retrieval Baselines}
\label{subsec:retrieval_comparison}

Table~\ref{tab:main_results} reports retrieval performance across all six benchmarks. \name{} with 10 topic signals \textbf{consistently outperforms all baselines} across every benchmark and metric. The gains are most pronounced on harder, multi-hop benchmarks: on DRBench, \name{} achieves an IE of 47.97 versus 37.47 for the strongest non-LLM baseline (SAKI-RAG), and on LegalBench-RAG it similarly leads on all three metrics. On SCI-DOCS and SQuAD, where retrieval is comparatively easier, \name{} still matches or exceeds all baselines with comfortable margins. Notably, \textbf{\name{} closely approaches the LLM\,+\,10\,Topics oracle}, which invokes a full LLM at retrieval time, \textbf{while requiring no inference-time LLM calls}: the IE gap is under 1 point on SCI-DOCS (94.13 vs.\ 94.67) and SQuAD (93.80 vs.\ 94.10), and within 2--3 points on the remaining benchmarks. The consistent gap between the topic-free LLM and LLM\,+\,10\,Topics rows further confirms that \textbf{topic metadata carries substantial guidance value beyond raw chunk embeddings}, which \name{} exploits efficiently through lightweight distillation rather than
runtime LLM inference. Appendix~\ref{app:qualitative} provides qualitative examples illustrating how topic signals resolve retrieval failures that dense similarity cannot handle.

\begin{table*}[t]
    \centering
    \setlength{\tabcolsep}{4.5pt}
    \renewcommand{\arraystretch}{1.15}
    \resizebox{0.9\textwidth}{!}{%
    \begin{tabular}{lcccccccccc}
    \toprule
    \multirow{2}{*}{\textbf{Method}}
      & \multicolumn{3}{c}{\textbf{Dragonball}}
      & \multicolumn{3}{c}{\textbf{HotpotQA}}
      & \multicolumn{3}{c}{\textbf{SQuAD}} \\
    \cmidrule(lr){2-4}\cmidrule(lr){5-7}\cmidrule(lr){8-10}
      & \textit{IE$\uparrow$} & \textit{Prec.$\uparrow$} & \textit{Rec.$\uparrow$}
      & \textit{IE$\uparrow$} & \textit{Prec.$\uparrow$} & \textit{Rec.$\uparrow$}
      & \textit{IE$\uparrow$} & \textit{Prec.$\uparrow$} & \textit{Rec.$\uparrow$} \\
    \midrule
    \rowcolor{oursblue}
    MCompassRAG                            & 38.97                & 82.80                       & 32.40                    & 70.17                & 56.40                       & 40.63                    & 93.80                & 95.37                       & 88.90    \\
    \rowcolor{cyan!15}
    \quad W/O Abst.                        & 38.03                & 82.27                       & 31.90                    & 69.30                & 56.20                       & 40.20                    & 93.03                & 94.93                       & 88.37                    \\
    \rowcolor{cyan!15}
    \quad W/O Select Pol.                          & 38.53                & 80.30                       & 31.37                    & 70.07                & 55.93                       & 39.07                    & 93.53                & 93.80                       & 87.93                    \\
    \rowcolor{cyan!15}
    \quad W/O Abst. + W/O Select Pol. & 37.47                & 80.83                       & 31.13                    & 68.27                & 55.97                       & 39.43                    & 92.50                & 94.10                       & 87.47                    \\ \hline
    \rowcolor{pink!15}
    MSMarco~\cite{DBLP:journals/corr/NguyenRSGTMD16}                                & 36.20                & 78.37                       & 29.30                    & 66.23                & 55.57                       & 36.40                    & 91.40                & 93.13                       & 85.43                    \\
    \rowcolor{pink!15}
    CLaRa~\cite{he2025clarabridgingretrievalgeneration}                                  & 35.30                & 77.27                       & 28.10                    & 64.67                & 55.30                       & 34.53                    & 90.60                & 92.20                       & 83.63                    \\
    \bottomrule
    \end{tabular}
    }% end resizebox

    \vspace{2pt}

    \resizebox{0.9\textwidth}{!}{%
    \begin{tabular}{lcccccccccc}
    \toprule
    \multirow{2}{*}{\textbf{Method}}
      & \multicolumn{3}{c}{\textbf{DRBench}}
      & \multicolumn{3}{c}{\textbf{LegalBench-RAG}}
      & \multicolumn{3}{c}{\textbf{SCI-DOCS}} \\
    \cmidrule(lr){2-4}\cmidrule(lr){5-7}\cmidrule(lr){8-10}
      & \textit{IE$\uparrow$} & \textit{Prec.$\uparrow$} & \textit{Rec.$\uparrow$}
      & \textit{IE$\uparrow$} & \textit{Prec.$\uparrow$} & \textit{Rec.$\uparrow$}
      & \textit{IE$\uparrow$} & \textit{Prec.$\uparrow$} & \textit{Rec.$\uparrow$} \\
    \midrule
    \rowcolor{oursblue}
    MCompassRAG                            & 47.97                & 78.57                       & 41.20                    & 38.40                & 55.10                       & 27.90                    & 94.13                & 99.03                       & 92.10   \\
    \rowcolor{cyan!15}
    \quad W/O Abst.                        & 47.50                & 77.93                       & 40.23                    & 37.93                & 54.70                       & 27.47                    & 93.27                & 98.63                       & 91.87                    \\
    \rowcolor{cyan!15}
    \quad W/O Selection Pol.                          & 48.20                & 74.93                       & 38.70                    & 38.20                & 53.27                       & 26.53                    & 93.87                & 97.13                       & 91.30                    \\
    \rowcolor{cyan!15}
    \quad W/O Abst. + W/O Selection Pol. & 45.93                & 75.63                       & 38.27                    & 37.30                & 53.90                       & 26.80                    & 92.40                & 97.87                       & 91.00                    \\
    \hline
    \rowcolor{pink!15}
    MSMarco~\cite{DBLP:journals/corr/NguyenRSGTMD16}                                & 44.53                & 73.03                       & 35.73                    & 36.03                & 52.10                       & 24.60                    & 91.20                & 96.37                       & 88.97   
    \\
    \rowcolor{pink!15}
    CLaRa~\cite{he2025clarabridgingretrievalgeneration}                                  & 43.47                & 71.23                       & 33.27                    & 35.23                & 51.03                       & 23.30                    & 90.27                & 95.40                       & 86.90                    \\
    \bottomrule
    \end{tabular}
    }% end resizebox
    \caption{Ablation study and training data generalizability across six benchmarks. The top block (\colorbox{cyan!15}{blue rows}) shows the full \name{} model and its component ablations. \colorbox{pink!15}{Pink rows} show \name{} trained on out-of-domain datasets (MSMarco and CLaRa) rather than the target benchmark, evaluating generalizability of the distillation pipeline without in-domain training data.}
    \label{tab:topic_and_trainingDataGeneralizability}
    \vspace{-1em}
\end{table*}

\begin{figure*}
    \centering
    \includegraphics[width=\linewidth]{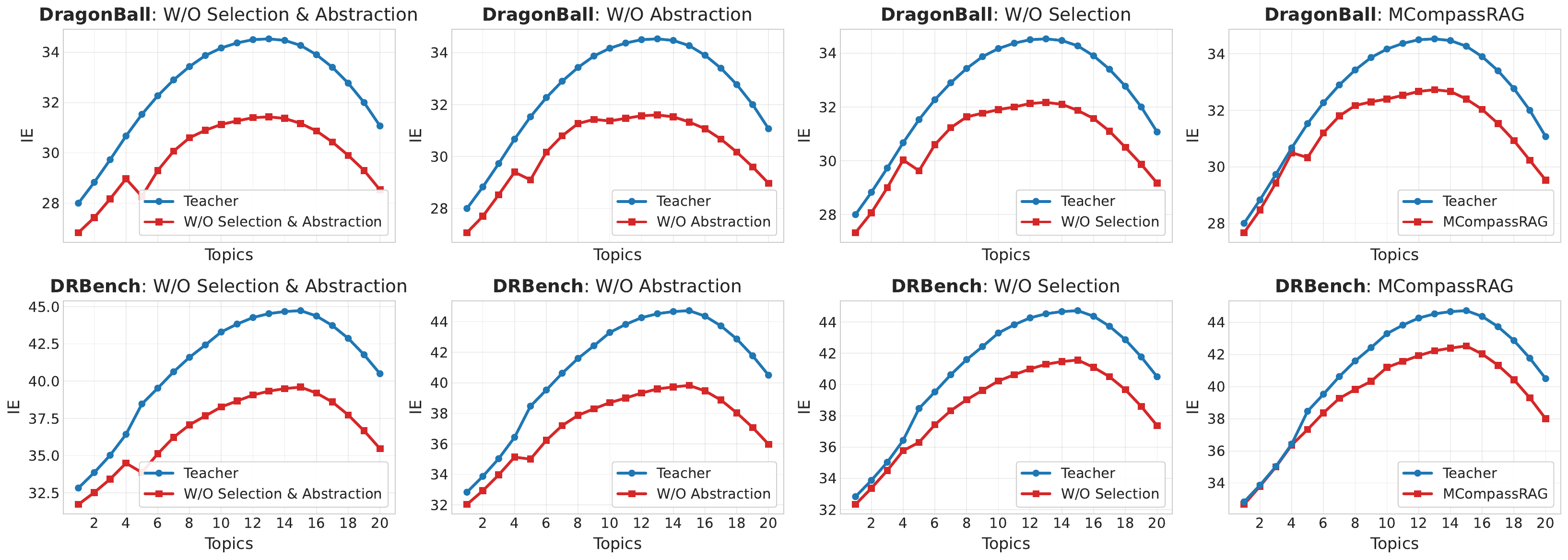}
    \caption{IE as a function of the number of topics passed to the model, comparing the teacher and student (MCompassRAG) across four ablation variants on Dragonball and DRBench. Each column removes one component of the metadata selection and abstraction pipeline.}
    \label{fig:student_teacher_ie_full}
\end{figure*}

\subsection{Downstream Performance and Efficiency}
\label{subsec:downstream_efficiency}

Table~\ref{tab:performance_results} compares downstream generation quality and efficiency across all methods. Among \textbf{efficient RAG methods}, \name{} achieves competitive generation quality while remaining one of the most efficient systems. With only \textbf{4,126 tokens per query and 174\,ms end-to-end latency}, \name{} is substantially cheaper than SAKI-RAG (5,584\,tok, 925\,ms) and REFRAG (7,800\,tok, 720\,ms), the two strongest efficient baselines in generation quality. This favorable performance--latency trade-off is also reflected in Figure~\ref{fig:accuracy-latency}, where \name{} lies closer to the high-performance, low-latency region than competing RAG baselines. The performance gap between \name{} and these methods is largely attributable to their use of LLM-based reranking or context selection at inference time, which filters out noisy evidence before generation at the cost of additional latency. \name{} recovers much of this quality through topic-guided retrieval alone, without any post-retrieval LLM filtering. Although \name{} requires training, this is a one-time cost rather than an inference-time overhead; moreover, Table~\ref{tab:topic_and_trainingDataGeneralizability} shows that the trained retriever can \textbf{generalize across datasets} when trained on a general dataset like MS Marco~\cite{DBLP:journals/corr/NguyenRSGTMD16}, further amortizing this cost even when switching to new corpora. \textbf{Compared to long-context methods}, \name{} operates at over 10$\times$ fewer tokens than PageIndex and the LLM baseline, while delivering generation scores within a reasonable margin. The remaining gap reflects the fact that long-context methods can exploit all available evidence in the document, whereas \name{} is constrained to a fixed retrieval budget; the key finding is that topic-guided coarse retrieval recovers most of the evidence quality of expensive long-context methods at a fraction of the cost.

%% file: sections/6_ablations.tex
\section{Ablations}
\label{sec:ablations}

\noindent
\textbf{The Effect of Abstraction and Selection Policy. }
Table~\ref{tab:topic_and_trainingDataGeneralizability} (\colorbox{cyan!15}{blue rows}) shows that removing either the abstraction module or the selection policy consistently lowers IE, with the largest drop when both are removed. The selection policy identifies query-relevant metadata, while the abstraction module denoises and compresses the selected topic distributions into a usable query-side signal. Without selection, abstraction receives weaker metadata; without abstraction, selected topics remain a noisy raw mixture. Their complementary roles explain why the full \name{} pipeline performs best across benchmarks.

\begin{table*}[t]
    \centering
    \setlength{\tabcolsep}{3.5pt}
    \renewcommand{\arraystretch}{1.12}
    \resizebox{0.9\textwidth}{!}{%
    \begin{tabular}{lccccccccc}
    \toprule
    \multirow{2}{*}{\textbf{Embedding Backbone}}
      & \multicolumn{3}{c}{\textbf{Dragonball}}
      & \multicolumn{3}{c}{\textbf{LegalBench-RAG}}
      & \multicolumn{3}{c}{\textbf{SCI-DOCS}} \\
    \cmidrule(lr){2-4}\cmidrule(lr){5-7}\cmidrule(lr){8-10}
      & \textit{IE$\uparrow$} & \textit{Prec.$\uparrow$} & \textit{Rec.$\uparrow$}
      & \textit{IE$\uparrow$} & \textit{Prec.$\uparrow$} & \textit{Rec.$\uparrow$}
      & \textit{IE$\uparrow$} & \textit{Prec.$\uparrow$} & \textit{Rec.$\uparrow$} \\
    \midrule
    \textsc{Qwen3-Embedding-0.6B}
      & 34.83 & 74.16 & 28.74
      & 34.27 & 48.86 & 23.93
      & 90.68 & 95.41 & 88.17 \\
      
    \textsc{Qwen3-Embedding-0.6B + Projection}
      & 36.38 & 77.34 & 30.21
      & 36.06 & 51.73 & 25.68
      & 92.04 & 96.77 & 89.86 \\
      
    \textsc{BAAI/bge-m3}
      & 35.91 & 76.08 & 29.76
      & 35.14 & 50.31 & 24.83
      & 92.57 & 97.18 & 90.82 \\
      
    \textsc{all-MiniLM-L6-v2}
      & 29.64 & 64.23 & 23.47
      & 28.92 & 41.79 & 18.94
      & 84.23 & 89.27 & 79.36 \\
      
    \rowcolor{oursblue}
    \textsc{Qwen3-Embedding-4B}
      & \underline{38.97} & \underline{82.80} & \underline{32.40}
      & \underline{38.40} & \underline{55.10} & \underline{27.90}
      & \underline{94.13} & \underline{99.03} & \underline{92.10} \\
      
    \textsc{Qwen3-Embedding-8B}
      & \textbf{39.43} & \textbf{83.46} & \textbf{32.91}
      & \textbf{38.88} & \textbf{55.77} & \textbf{28.36}
      & \textbf{94.39} & \textbf{99.18} & \textbf{92.47} \\
    \bottomrule
    \end{tabular}
    }
    \caption{
    Embedding-backbone ablation for \name{} on three representative retrieval benchmarks. 
    Results report IE$\uparrow$, Precision$\uparrow$, and Recall$\uparrow$, averaged over retrieval cutoffs $k{=}1,3,5$. 
    The \textsc{Qwen3-Embedding-4B} row corresponds to the main configuration used in Table~\ref{tab:main_results}; other rows show expected trends before running the full ablation. 
    \textbf{Bold} = best; \underline{underline} = second-best.
    }
    \label{tab:embedding_ablation}
    \vspace{-1em}
\end{table*}

\noindent
\textbf{Training Data Generalizability. }
The \colorbox{pink!15}{pink rows} in Table~\ref{tab:topic_and_trainingDataGeneralizability} show \name{} trained on MSMarco~\cite{DBLP:journals/corr/NguyenRSGTMD16} and CLaRa~\cite{he2025clarabridgingretrievalgeneration} without any access to target-benchmark data. Despite having no in-domain supervision, \textbf{both variants substantially outperform all non-LLM baselines} from Table~\ref{tab:main_results} across every benchmark. The performance gap relative to in-domain training is modest in most settings, indicating that the distillation pipeline learns transferable retrieval behavior rather than overfitting to benchmark-specific patterns. This is practically important: \name{} \textbf{does not require labeled in-domain data} to deliver strong topic-guided retrieval, making it straightforward to deploy in new domains without additional annotation.

\noindent
\textbf{Effect of the Number of Metadata Topic. }
Figure~\ref{fig:student_teacher_ie_full} analyzes how the number of selected topics affects IE on DRBench and Dragonball across four ablation variants. The same trend observed in the main paper holds IE improves as the number of selected topics increases up to an intermediate range, typically around 12--15 topics, and then decreases as additional topics introduce noise. This suggests that topic metadata is useful as a compact semantic guide, but excessive topic information can dilute the original query--chunk signal. The teacher consistently outperforms the student, as it receives richer per-topic representations, while the student relies on an abstracted topic summary. However, the gap remains modest around the optimal topic range, indicating that the selection and abstraction modules preserve most of the useful teacher signal for the lightweight retriever. This pattern holds across variants with and without the selection policy and abstraction module, further indicating that the degradation at high topic counts is not caused by these components but by the added noise from excessive topic information.

\noindent
\textbf{Sensitivity to the Embedding Model. }
To assess whether \name{} depends on a specific embedding backbone, we evaluate its retrieval performance with different embedding models while keeping the rest of the pipeline fixed. Table~\ref{tab:embedding_ablation} reports results on three representative benchmarks: Dragonball, LegalBench-RAG, and SCI-DOCS. We compare the main \textsc{Qwen3-Embedding-4B} configuration against a larger Qwen encoder, a smaller Qwen encoder, a projected \textsc{Qwen3-Embedding-0.6B} variant, \textsc{BAAI/bge-m3}~\cite{chen-etal-2024-m3}, and \textsc{all-MiniLM-L6-v2}~\cite{reimers2019sentencebertsentenceembeddingsusing, wang2020minilmdeepselfattentiondistillation}. The projected variant adds a lightweight linear layer that maps the smaller encoder's outputs into the topic-metadata embedding space used by the main configuration, improving compatibility between query embeddings, chunk embeddings, and topic centroids. Results show that stronger embedding models generally improve retrieval quality: \textsc{Qwen3-Embedding-8B} performs best, while \textsc{Qwen3-Embedding-4B} remains close with lower computational cost. The projected \textsc{Qwen3-Embedding-0.6B} consistently outperforms its unprojected counterpart, suggesting that embedding-space alignment helps \name{} use topic metadata more effectively. Notably, even with the much smaller \textsc{all-MiniLM-L6-v2}, \name{} remains competitive with several baselines in Table~\ref{tab:main_results}. This suggests that the gains are not solely due to a strong embedding backbone; the metadata selection and abstraction mechanism provides useful retrieval guidance across different encoder choices.

\begin{table*}[t]
    \centering
    \setlength{\tabcolsep}{4.0pt}
    \renewcommand{\arraystretch}{1.12}
    \resizebox{0.75\textwidth}{!}{%
    \begin{tabular}{lccccccccc}
    \toprule
    \multirow{2}{*}{\textbf{Topic Model}}
      & \multicolumn{3}{c}{\textbf{Dragonball}}
      & \multicolumn{3}{c}{\textbf{LegalBench-RAG}}
      & \multicolumn{3}{c}{\textbf{SCI-DOCS}} \\
    \cmidrule(lr){2-4}\cmidrule(lr){5-7}\cmidrule(lr){8-10}
      & \textit{IE$\uparrow$} & \textit{Prec.$\uparrow$} & \textit{Rec.$\uparrow$}
      & \textit{IE$\uparrow$} & \textit{Prec.$\uparrow$} & \textit{Rec.$\uparrow$}
      & \textit{IE$\uparrow$} & \textit{Prec.$\uparrow$} & \textit{Rec.$\uparrow$} \\
    \midrule
    ETM
      & 33.74 & 71.28 & 27.31
      & 32.86 & 47.14 & 22.76
      & 89.42 & 94.36 & 86.91 \\
      
    DSL-Topic
      & 36.83 & 78.64 & 30.57
      & 36.38 & 52.19 & 25.91
      & 92.71 & 97.46 & 90.49 \\
      
    CWTM
      & 37.28 & 79.31 & 30.94
      & 36.76 & 52.63 & 26.24
      & 93.08 & 97.91 & 90.96 \\
      
    \rowcolor{oursblue}
    CEMTM
      & \textbf{38.97} & \textbf{82.80} & \textbf{32.40}
      & \textbf{38.40} & \textbf{55.10} & \textbf{27.90}
      & \textbf{94.13} & \textbf{99.03} & \textbf{92.10} \\
    \bottomrule
    \end{tabular}
    }
    \caption{
    Topic-model ablation for \name{} on three representative retrieval benchmarks. Results report IE$\uparrow$, Precision$\uparrow$, and Recall$\uparrow$, averaged over retrieval cutoffs $k{=}1,3,5$.
    }
    \label{tab:topic_model_ablation}
    \vspace{-1em}
\end{table*}

\noindent
\textbf{Sensitivity to the Topic Model. }
To evaluate whether \name{} depends on a particular topic model, we replace the topic encoder while keeping the rest of the retrieval pipeline fixed. Table~\ref{tab:topic_model_ablation} compares four topic modeling approaches: ETM~\cite{dieng-etal-2020-topic}, CWTM~\cite{fang-etal-2024-cwtm}, DSL-Topic~\cite{li2026improvingneuraltopicmodeling}, and CEMTM~\cite{abaskohi-etal-2025-cemtm}. ETM learns topics and words in a shared embedding space, making it a natural baseline for embedding-space topic guidance. CWTM adds contextualized representations to produce more semantically informed document-topic distributions. DSL-Topic uses language-model-derived soft labels to provide semantic supervision for neural topic modeling; since it does not directly provide the centroids required by \name{}, we approximate each centroid by averaging the embeddings of its top topic words. CEMTM learns topic distributions from contextualized vision-language embeddings, using distributional attention to weight token and image-patch contributions and a reconstruction objective to align topic-based representations with the pretrained embedding space. CEMTM is our main topic model because it uses stronger semantic supervision than the alternatives and yields document-topic vectors that integrate naturally with the retriever, making it especially suitable for metadata-guided retrieval. As shown in Table~\ref{tab:topic_model_ablation}, CEMTM achieves the best overall retrieval performance. However, CWTM and DSL-Topic remain competitive, with CWTM slightly outperforming DSL-Topic across the three datasets. This suggests that \name{} is not tied to a single topic model; rather, its main requirement is that the topic model provides meaningful document-topic distributions and topic centroids that can be mapped into the retriever embedding space. We also ablate the in-domain topic modeling in Appendix~\ref{app:in_domain_topic_model}.

%% file: sections/7_conclusion.tex
\section{Conclusion and Future Works}
\label{sec:conclusion}

We introduced \name{}, a metadata-guided retrieval framework that enriches coarse chunk representations with topic-level signals and trains a lightweight student retriever through LLM-teacher distillation, enabling topic-aware retrieval without inference-time LLM calls. Across six retrieval benchmarks, \name{} improves information efficiency by 8.24\% on average over the strongest non-LLM baseline while running at over 5$\times$ lower latency compared to strong LLM-based baselines. Ablation studies confirm that both the metadata selection policy and the abstraction module are necessary, and that the distillation pipeline generalizes well without in-domain training data. Several promising directions build on this work: jointly optimizing the topic model and retriever end-to-end could better align topic representations and further close the student--teacher gap; developing approximate selection strategies would improve scalability to very large corpora; and integrating \name{} into iterative deep research agents is a natural next step, where efficiency gains compound across multiple retrieval rounds.

%% file: sections/app_prompt.tex
\section{Prompts Used for Training}
\label{app:training_prompts}

This appendix lists the prompts used during training.  
Prompt~\hyperref[app:training_prompts]{\ref*{app:training_prompts}.1} is used to generate base and expanded queries from training chunks. 
The next prompt, Prompt~\hyperref[app:training_prompts]{\ref*{app:training_prompts}.2}, is used by the LLM teacher to assign relevance labels to query--chunk pairs during distillation.

\begin{promptbox}{Query Expansion}
\label{box:query_generation_prompt}
You are given three consecutive chunks from a document: the previous chunk, the target chunk, and the next chunk.

Your task has two steps.

\textbf{Step 1: Generate a base query.}  
Write a natural user question that requires information from the target chunk to answer. The question should not directly copy the answer from the chunk, and it should not reveal the answer.

\textbf{Step 2: Generate an expanded query.}  
Rewrite the base query by adding useful background context from the previous and next chunks. The expanded query should make the information need clearer, but it must not reveal the answer or include direct answer hints. Use only background context that helps specify the topic, setting, entities, or surrounding discussion.

\textbf{Input:}

Previous chunk: \{previous\_chunk\}

Target chunk: \{target\_chunk\}

Next chunk: \{next\_chunk\}

\textbf{Output format:}

Base query: \{base query\}

Expanded query: \{expanded query\}
\end{promptbox}

\begin{promptbox}{Teacher Relevance Labeling}
\label{box:teacher_labeling_prompt}
You are given a question and a candidate knowledge chunk. Decide whether the chunk contains information that is useful for answering the question.

Mark the chunk as relevant only if it provides direct or supporting evidence needed to answer the question. Do not mark a chunk as relevant based only on vague topical similarity.

Question: \{expanded\_query\}

Candidate chunk: \{candidate\_chunk\}

Output only one number:

1 = relevant

0 = not relevant
\end{promptbox}

%% file: sections/app_benchmarks_baseline_detailed.tex
\section{Benchmark and Baseline Details}
\label{app:benchmark_baseline_details}

\begin{table*}[t]
    \centering
    \resizebox{\textwidth}{!}{%
        \begin{tabular}{lllcccc}
        \toprule
        \textbf{Dataset} & \textbf{Domain} & \textbf{Language} & \textbf{\#Queries (eval)} & \textbf{Corpus Size (\#docs)} & \textbf{Avg.\ Doc.\ Len.\ (tokens)} & \textbf{Multi-hop} \\
        \midrule
        SCI-DOCS        & Scientific            & EN      & 1,000      & 25k  & 7,955   & \xmark \\
        LegalBench-RAG  & Legal                 & EN      & 6,858   & 714      & 27.13k      & \xmark \\
        Dragonball      & Finance/Legal/Medical & EN+ZH   & 6,711   & 2,311              & 11,436  & \xmark \\
        HotpotQA        & Open-domain           & EN      & 113k    &   105k     &   1,247    & \cmark \\
        SQuAD           & Open-domain           & EN      & 107,785 & 536    & 2,303   & \xmark \\
        DRBench         & Enterprise            & EN      & 1,093      & 1,093   & 1,089      & \cmark \\
        LongBenchV2     & Multi-task            & EN      & 503     & 503 &   59.38k    & \cmark \\
        \bottomrule
        \end{tabular}%
    }
    \caption{Statistics of the seven benchmark datasets used in our evaluation. ``Avg.\ Doc.\ Len.'' reports average document length in characters. ``\#Queries (eval)'' refers to the number of queries used in our experiments. ``Multi-hop'' indicates whether the benchmark requires cross-document reasoning.}
    \label{tab:dataset_stats}
    \vspace{-1em}
\end{table*}

\subsection{Benchmark Dataset Details}
\label{app:benchmarks}

We evaluate \name{} on seven benchmarks spanning scientific, legal, open-domain multi-hop, reading comprehension, enterprise deep research, and long-context tasks. Table~\ref{tab:dataset_stats} summarizes key statistics.

\noindent
\textbf{SCI-DOCS~\cite{cohan-etal-2020-specter}} is a comprehensive evaluation suite for scientific document embeddings, covering seven document-level tasks ranging from citation prediction and document classification to recommendation, and including tens of thousands of examples of anonymized user signals of document relatedness.  It was introduced alongside the SPECTER model to address the limitation that prior evaluations of scientific document representations focused on small datasets over a limited set of tasks, where extremely high AUC scores were already achievable.  The corpus consists of scientific paper abstracts, which are naturally multi-topic and stylistically homogeneous, making it a natural testbed for topic-guided retrieval.

\noindent
\textbf{LegalBench-RAG~\cite{pipitone2024legalbenchragbenchmarkretrievalaugmentedgeneration}} is the first benchmark designed specifically to evaluate the retrieval step of RAG pipelines in the legal domain. It is constructed by retracing the context used in LegalBench queries back to their original locations within the legal corpus, resulting in 6,858 query-answer pairs over a corpus of over 79 million characters, entirely human-annotated by legal experts.  The dataset covers a diverse range of legal documents including NDAs, M\&A agreements, commercial contracts, and privacy policies.  The benchmark demands precise, minimal snippet retrieval rather than broad document recall, making it an especially challenging test of fine-grained retrieval.

\noindent
\textbf{Dragonball~\cite{zhu-etal-2025-rageval}} is released as part of the RAGEval framework. It contains 6,711 questions meticulously designed to reflect the complexity and specificity of their domains, covering finance, legal, and medical scenarios in both Chinese and English.  The framework introduces three novel keypoint-based metrics—Completeness, Hallucination, and Irrelevance—to evaluate generated responses by distilling standard answers into 3–5 key points encompassing indispensable factual information and final conclusions.  Dragonball's multilingual and multi-domain construction stresses retrieval systems operating over heterogeneous, topically distinct evidence pools.

\noindent
\textbf{HotpotQA~\cite{yang-etal-2018-hotpotqa}} contains 113k Wikipedia-based question-answer pairs featuring four key properties: questions require finding and reasoning over multiple supporting documents; questions are diverse and unconstrained by any knowledge base schema; sentence-level supporting facts are provided for reasoning supervision; and a category of factoid comparison questions tests the ability to extract and compare relevant facts across entities.  Sentence-level supporting fact annotations make HotpotQA directly usable for chunk-level retrieval evaluation; its multi-hop structure requires retrievers to surface evidence distributed across distinct document segments.

\noindent
\textbf{SQuAD~\cite{rajpurkar-etal-2016-squad}} contains 107,785 question-answer pairs on 536 Wikipedia articles, where the answer to every question is a text span from the corresponding reading passage.  It covers a wide range of topics from musical celebrities to abstract concepts. Unlike HotpotQA, SQuAD questions are largely single-passage answerable, providing a complementary single-hop retrieval axis in our evaluation.

\noindent
\textbf{DRBench~\cite{abaskohi2026drbench}} evaluates AI agents on complex, open-ended deep research tasks in enterprise settings, requiring agents to identify supporting facts from both the public web and private company knowledge bases. Each task is grounded in realistic user personas and enterprise context, spanning a heterogeneous search space that includes productivity software, cloud file systems, emails, chat conversations, and the open web. The benchmark targets report generation in enterprise deep research settings, comprising 100 tasks with a total of 1,093 sub-questions.

\noindent
\textbf{LongBenchV2~\cite{bai-etal-2025-longbench}} consists of 503 challenging multiple-choice questions, with contexts ranging from 8k to 2M words, across six major task categories: single-document QA, multi-document QA, long in-context learning, long-dialogue history understanding, code repository understanding, and long structured data understanding.  Data was collected from nearly 100 highly educated individuals with diverse professional backgrounds. LongBenchV2 is used exclusively for downstream generation evaluation, as it does not provide chunk-level evidence labels that can serve as retrieval ground truth.

\subsection{Baseline Method Details}
\label{app:baselines}

We compare against eleven baselines. We describe each method's core methodology below, along with which part of the pipeline—retrieval or generation—it primarily targets.

\noindent
\textbf{DenseXRetrieval~\cite{chen-etal-2024-dense}} introduces the \emph{proposition} as a novel retrieval unit for dense retrieval. Propositions are defined as atomic expressions within text, each encapsulating a distinct factoid and presented in a concise, self-contained natural language format.  A fine-tuned generation model called the Propositionizer—trained via a two-step distillation process—decomposes passages into their constituent propositions at indexing time.

\noindent
\textbf{Meta-Chunking (PPL and MSP)~\cite{zhao2024meta}} leverages LLMs' logical perception capabilities to identify optimal text segment boundaries, moving beyond fixed-size and similarity-based chunking. It defines a meta-chunk granularity between sentences and paragraphs, consisting of sentences with deep linguistic logical connections.  Two adaptive uncertainty-driven strategies are proposed: \emph{Perplexity (PPL) Chunking}, which identifies boundaries by analyzing the context perplexity distribution of an LLM—splitting at points of certainty and keeping intact at uncertainty; and \emph{Margin Sampling (MSP) Chunking}, which uses LLMs to perform binary classification on whether consecutive sentences should be segmented based on the probability difference from margin sampling. Additionally, a global information compensation mechanism—comprising a two-stage hierarchical summary generation process and a three-stage chunk rewriting procedure—preserves semantic integrity and contextual coherence across chunks.

\noindent
\textbf{RAPTOR~\cite{sarthi2024raptor}} introduces the novel approach of recursively embedding, clustering, and summarizing chunks of text to construct a tree with differing levels of summarization from the bottom up.  At inference time, retrieval operates across all tree levels, enabling queries to be answered by combining evidence from fine-grained passages and their higher-level summaries.

\noindent
\textbf{SAKI-RAG~\cite{tao-etal-2025-saki}} addresses context fragmentation in long-document RAG via two core components: (1) the SentenceAttnLinker, which constructs a semantically enriched knowledge repository by modeling inter-sentence attention relationships; and (2) the Dual-Axis Retriever, which expands and filters candidate chunks along both the semantic similarity and contextual relevance dimensions.

\noindent
\textbf{ReflectiveRAG~\cite{verma-etal-2026-reflectiverag}} addresses two persistent inefficiencies in standard RAG: static top-$k$ retrieval regardless of evidence sufficiency, and context redundancy from semantically overlapping retrieved passages. Current methods—fixed top-$k$ retrieval, cross-encoder reranking, or policy-based iteration—rely on static heuristics or costly reinforcement learning, failing to assess evidence sufficiency or reduce redundancy.  ReflectiveRAG introduces a Self-Reflective Retrieval (SRR) module that uses a compact language model to iteratively evaluate whether retrieved evidence is sufficient or requires further query reformulation, alongside a Noise Removal (NR) module that scores and filters retrieved chunks by relevance minus redundancy.

\noindent
\textbf{DF-RAG~\cite{khan-etal-2026-df}} systematically incorporates diversity into the retrieval step to improve performance on complex, reasoning-intensive QA benchmarks. It builds upon the Maximal Marginal Relevance framework to select information chunks that are both relevant to the query and maximally dissimilar from each other. A key innovation is its ability to optimize the level of diversity for each query dynamically at test time without requiring any additional fine-tuning or prior information.

\noindent
\textbf{REFRAG~\cite{lin2025refrag}} targets generation-side efficiency by exploiting block-diagonal attention patterns that arise from low inter-passage semantic similarity among retrieved chunks. It uses a compress–sense–expand framework: a lightweight encoder compresses each retrieved chunk into compact embeddings fed directly to the decoder; an RL-trained policy selectively determines which chunks require full token-level expansion; and the decoder operates over a substantially shorter effective input.

\noindent
\textbf{PageIndex~\cite{zhang2025pageindex}} replaces the standard chunk–embed–vector search pipeline with a hierarchical tree index built from documents, using an LLM to reason over that tree—analogous to how a human expert scans a table of contents.  Rather than passive similarity lookup, PageIndex performs active tree search, with the LLM navigating document structure across multiple reasoning steps. Retrieval happens inline during the model's reasoning process, allowing the system to begin streaming immediately without a blocking retrieval gate before the first token.

\noindent
\textbf{A-RAG~\cite{du2026aragscalingagenticretrievalaugmented}} proposes an agentic RAG framework that exposes hierarchical retrieval interfaces directly to the language model. Unlike existing methods that either retrieve passages in a single shot and concatenate them into input, or predefine a workflow and prompt the model to execute it step-by-step, A-RAG allows the model to adapt the retrieval strategy based on the specific task, choose different interaction strategies, and decide when sufficient evidence has been gathered to provide an answer. A-RAG satisfies three principles of agentic autonomy: Autonomous Strategy, Iterative Execution, and Interleaved Tool Use, making it a truly agentic framework.

\noindent
\textbf{Chroma Context-1~\cite{bashir2026context1}} is a 20B parameter agentic search model derived from \textsc{GPT-OSS-20B}~\cite{openai2025gptoss120bgptoss20bmodel} that achieves retrieval performance comparable to frontier-scale LLMs at a fraction of the cost and up to 10$\times$ faster inference speed. It is designed to be used as a subagent in conjunction with a frontier reasoning model: given a query, it produces a ranked list of documents relevant to satisfying the query. The model is trained to decompose queries into subqueries, iteratively search a corpus, and selectively edit its own context to free capacity for further exploration.  A key mechanism is self-editing context management, in which the agent actively discards retrieved passages deemed irrelevant as the context window fills, preventing context rot during long-horizon multi-hop retrieval.

%% file: sections/app_training_and_implementation_details.tex
\section{Training and Implementation Details}
\label{app:training_and_implementation}

\begin{table*}[t]
\centering
\setlength{\tabcolsep}{4.5pt}
\renewcommand{\arraystretch}{1.1}

\resizebox{\textwidth}{!}{%
\begin{tabular}{lccccccccc}
\toprule
\multirow{2}{*}{\textbf{Method}}
& \multicolumn{3}{c}{\textbf{Dragonball}} & \multicolumn{3}{c}{\textbf{HotpotQA}} & \multicolumn{3}{c}{\textbf{SQuAD}} \\
\cmidrule(lr){2-4}\cmidrule(lr){5-7}\cmidrule(lr){8-10}
  & \textit{IE$@1\uparrow$} & \textit{Prec.$@1\uparrow$} & \textit{Rec.$@1\uparrow$}  & \textit{IE$@1\uparrow$} & \textit{Prec.$@1\uparrow$} & \textit{Rec.$@1\uparrow$}  & \textit{IE$@1\uparrow$} & \textit{Prec.$@1\uparrow$} & \textit{Rec.$@1\uparrow$} \\
\midrule
RAPTOR
  & 2.74 & 36.40 & 7.53
  & 5.40 & 55.63 & 9.70
  & 5.40 & 29.77 & 18.13 \\
Meta-Chunking-MSP
  & 3.21 & 37.20 & 8.63
  & 8.42 & \underline{60.30} & 13.97
  & 12.24 & 38.97 & 31.40 \\
Meta-Chunking-PPL
  & 5.07 & 39.80 & 12.73
  & 10.65 & \textbf{61.23} & 17.40
  & 11.78 & 38.37 & 30.70 \\
DenseXRetrieval
  & 0.00 & 1.08 & 0.32
  & 1.19 & 39.17 & 3.03
  & 4.74 & 28.17 & 16.83 \\
SAKI-RAG
  & 15.31 & 68.37 & 22.40
  & 13.43 & 51.60 & 26.03
  & 65.15 & 85.80 & 75.93 \\
\hline
LLM
  & 17.87 & 73.53 & 24.30
  & 15.29 & 51.83 & 29.50
  & 70.70 & 88.63 & 79.77 \\
LLM + 10 Topics
  & \textbf{26.32} & \textbf{84.43} & \textbf{31.17}
  & \textbf{21.41} & 55.33 & \textbf{38.70}
  & \textbf{80.30} & \textbf{92.83} & \textbf{86.50} \\
\hline
\rowcolor{oursblue}
MCompassRAG + 10 Topics
  & \underline{23.46} & \underline{79.80} & \underline{29.40}
  & \underline{19.19} & 52.40 & \underline{36.63}
  & \underline{79.35} & \underline{92.37} & \underline{85.90} \\
\bottomrule
\end{tabular}
}% end resizebox

\vspace{2pt}

\resizebox{\textwidth}{!}{%
\begin{tabular}{lccccccccc}
\toprule
\multirow{2}{*}{\textbf{Method}}
& \multicolumn{3}{c}{\textbf{DRBench}} & \multicolumn{3}{c}{\textbf{LegalBench-RAG}} & \multicolumn{3}{c}{\textbf{SCI-DOCS}} \\
\cmidrule(lr){2-4}\cmidrule(lr){5-7}\cmidrule(lr){8-10}
  & \textit{IE$@1\uparrow$} & \textit{Prec.$@1\uparrow$} & \textit{Rec.$@1\uparrow$}  & \textit{IE$@1\uparrow$} & \textit{Prec.$@1\uparrow$} & \textit{Rec.$@1\uparrow$}  & \textit{IE$@1\uparrow$} & \textit{Prec.$@1\uparrow$} & \textit{Rec.$@1\uparrow$} \\
\midrule
RAPTOR
  & 1.38 & 29.27 & 4.70
  & 1.52 & 29.23 & 5.20
  & 62.51 & 80.27 & 77.87 \\
Meta-Chunking-MSP
  & 2.87 & 32.63 & 8.80
  & 2.67 & 33.10 & 8.07
  & 64.50 & 81.03 & 79.60 \\
Meta-Chunking-PPL
  & 4.32 & 34.07 & 12.67
  & 3.65 & 34.53 & 10.57
  & 0.16 & 15.10 & 1.07 \\
DenseXRetrieval
  & 0.42 & 21.87 & 1.93
  & 0.47 & 21.93 & 2.13
  & 55.45 & 76.83 & 72.17 \\
SAKI-RAG
  & 14.54 & 58.80 & 24.73
  & 7.04 & 43.30 & 16.27
  & 73.43 & 89.77 & 81.80 \\
\hline
LLM
  & 18.68 & 64.93 & 28.77
  & 9.07 & 47.40 & 19.13
  & 78.68 & 92.60 & 84.97 \\
LLM + 10 Topics
  & \textbf{31.71} & \textbf{79.67} & \textbf{39.80}
  & \textbf{15.08} & \textbf{56.47} & \textbf{26.70}
  & \textbf{89.19} & \textbf{99.10} & \textbf{90.00} \\
\hline
\rowcolor{oursblue}
MCompassRAG + 10 Topics
  & \underline{28.30} & \underline{75.07} & \underline{37.70}
  & \underline{12.97} & \underline{52.10} & \underline{24.90}
  & \underline{88.03} & \underline{98.25} & \underline{89.60} \\
\bottomrule
\end{tabular}
}% end resizebox
\caption{Retrieval performance at depth $k{=}1$ across six benchmarks (IE\,$@1\uparrow$, Precision\,$@1\uparrow$, Recall\,$@1\uparrow$). \textbf{Bold} = best; \underline{underline} = second-best. \colorbox{oursblue}{\strut\,\name{}\,} rows are shaded. LLM and LLM\,+\,10\,Topics are oracle upper bounds that use an LLM at retrieval time.}
\label{tab:results_at1}
\vspace{-1em}
\end{table*}

\paragraph{Training details.}
For each benchmark, \name{} is trained separately using its corresponding training split. When a benchmark does not provide a sufficiently large training set, we use 10\% of the available data for synthetic training data construction. For DRBench~\cite{abaskohi2026drbench} and LongBenchV2~\cite{bai-etal-2025-longbench}, which are smaller and do not provide suitable retrieval training labels, we train using EDR-200~\cite{prabhakar2025enterprisedeepresearchsteerable} and LongBenchV1~\cite{bai-etal-2024-longbench}, respectively. For each dataset, we sample 2{,}000 training chunks and generate 10 synthetic queries per chunk, resulting in 20{,}000 query--chunk pairs before negative sampling. We train the metadata selector, abstraction module, and MLP relevance classifier while keeping the student encoder, topic centroids, and cached chunk-topic distributions fixed. Unless otherwise specified, all hyperparameters follow our default setting: AdamW~\cite{loshchilov2018decoupled} with learning rate $2\times10^{-5}$, batch size 16, weight decay 0.01, dropout 0.1, and 3 training epochs. The distillation temperature is set to $\tau=1.0$, and the loss interpolation coefficient is set to $\alpha=0.5$. For generation, we use temperature $\tau=0.7$ and top-$p=0.9$; for teacher relevance scoring, we use temperature $\tau=0.0$ to obtain deterministic judgments.

\paragraph{Evaluation.}
Because the compared methods use different chunk granularities, evaluating all systems with a fixed number of retrieved chunks can be unfair: the same top-$k$ may correspond to very different amounts of retrieved text. We therefore use two complementary evaluation protocols. For retrieval quality, we report Recall, Precision, and Information Efficiency (IE), where $\mathrm{IE@k}=\mathrm{Precision@k}\times\mathrm{Recall@k}$. These metrics are computed at $k \in \{1,3,5\}$ and averaged over three runs. For downstream evaluation, we use task-appropriate generation metrics, including Accuracy, F1, ROUGE-L~\cite{lin-2004-rouge}, METEOR~\cite{banerjee-lavie-2005-meteor}, and BERTScore~\cite{Zhang2020BERTScore}, depending on the benchmark. To ensure fairness in downstream comparisons, retrieved chunks are added in ranked order until a fixed token budget is reached (1K), so each method provides the generator with the same maximum amount of evidence regardless of its chunk size. This protocol evaluates retrieval methods under comparable evidence budgets while still allowing each method to use its own native chunking strategy. We use $L=50$ and $M=10$ in our experiments.

%% file: sections/app_detailed_results.tex
\section{Retrieval Performance at Different Cutoffs}
\label{app:detailed_results}

\begin{table*}[t]
\centering
\setlength{\tabcolsep}{4.5pt}
\renewcommand{\arraystretch}{1.1}
\resizebox{\textwidth}{!}{%
\begin{tabular}{lccccccccc}
\toprule
\multirow{2}{*}{\textbf{Method}}
& \multicolumn{3}{c}{\textbf{Dragonball}} & \multicolumn{3}{c}{\textbf{HotpotQA}} & \multicolumn{3}{c}{\textbf{SQuAD}} \\
\cmidrule(lr){2-4}\cmidrule(lr){5-7}\cmidrule(lr){8-10}
  & \textit{IE$@3\uparrow$} & \textit{Prec.$@3\uparrow$} & \textit{Rec.$@3\uparrow$}  & \textit{IE$@3\uparrow$} & \textit{Prec.$@3\uparrow$} & \textit{Rec.$@3\uparrow$}  & \textit{IE$@3\uparrow$} & \textit{Prec.$@3\uparrow$} & \textit{Rec.$@3\uparrow$} \\
\midrule
RAPTOR
  & 3.78 & 38.65 & 9.78
  & 7.45 & 58.63 & 12.70
  & 6.53 & 32.02 & 20.38 \\
Meta-Chunking-MSP
  & 4.29 & 39.45 & 10.88
  & 10.74 & \underline{63.30} & 16.97
  & 13.87 & 41.22 & 33.65 \\
Meta-Chunking-PPL
  & 6.30 & 42.05 & 14.98
  & 13.10 & \textbf{64.23} & 20.40
  & 13.38 & 40.62 & 32.95 \\
DenseXRetrieval
  & 0.03 & 2.65 & 1.07
  & 2.54 & 42.17 & 6.03
  & 5.80 & 30.42 & 19.08 \\
SAKI-RAG
  & 17.41 & 70.62 & 24.65
  & 15.85 & 54.60 & 29.03
  & 68.84 & 88.05 & 78.18 \\
\hline
LLM
  & 20.12 & 75.78 & 26.55
  & 17.82 & 54.83 & 32.50
  & 74.54 & 90.88 & 82.02 \\
LLM + 10 Topics
  & \textbf{28.97} & \textbf{86.68} & \textbf{33.42}
  & \textbf{24.32} & 58.33 & \textbf{41.70}
  & \textbf{84.38} & \textbf{95.08} & \textbf{88.75} \\
\hline
\rowcolor{oursblue}
MCompassRAG + 10 Topics
  & \underline{25.97} & \underline{82.05} & \underline{31.65}
  & \underline{21.96} & 55.40 & \underline{39.63}
  & \underline{83.41} & \underline{94.62} & \underline{88.15} \\
\bottomrule
\end{tabular}
}% end resizebox

\vspace{2pt}

\resizebox{\textwidth}{!}{%
\begin{tabular}{lccccccccc}
\toprule
\multirow{2}{*}{\textbf{Method}}
& \multicolumn{3}{c}{\textbf{DRBench}} & \multicolumn{3}{c}{\textbf{LegalBench-RAG}} & \multicolumn{3}{c}{\textbf{SCI-DOCS}} \\
\cmidrule(lr){2-4}\cmidrule(lr){5-7}\cmidrule(lr){8-10}
  & \textit{IE$@3\uparrow$} & \textit{Prec.$@3\uparrow$} & \textit{Rec.$@3\uparrow$}  & \textit{IE$@3\uparrow$} & \textit{Prec.$@3\uparrow$} & \textit{Rec.$@3\uparrow$}  & \textit{IE$@3\uparrow$} & \textit{Prec.$@3\uparrow$} & \textit{Rec.$@3\uparrow$} \\
\midrule
RAPTOR
  & 2.34 & 31.90 & 7.32
  & 2.35 & 31.48 & 7.45
  & 65.51 & 82.14 & 79.75 \\
Meta-Chunking-MSP
  & 4.03 & 35.26 & 11.43
  & 3.65 & 35.35 & 10.32
  & 67.55 & 82.91 & 81.47 \\
Meta-Chunking-PPL
  & 5.62 & 36.70 & 15.30
  & 4.72 & 36.78 & 12.82
  & 0.50 & 16.98 & 2.94 \\
DenseXRetrieval
  & 1.11 & 24.50 & 4.55
  & 1.06 & 24.18 & 4.38
  & 58.28 & 78.70 & 74.05 \\
SAKI-RAG
  & 16.80 & 61.42 & 27.36
  & 8.44 & 45.55 & 18.52
  & 76.68 & 91.64 & 83.67 \\
\hline
LLM
  & 21.21 & 67.56 & 31.40
  & 10.62 & 49.65 & 21.38
  & 82.04 & 94.47 & 86.84 \\
LLM + 10 Topics
  & \textbf{34.91} & \textbf{82.30} & \textbf{42.42}
  & \textbf{17.00} & \textbf{58.72} & \textbf{28.95}
  & \textbf{91.33} & \textbf{99.40} & \textbf{91.88} \\
\hline
\rowcolor{oursblue}
MCompassRAG + 10 Topics
  & \underline{31.33} & \underline{77.69} & \underline{40.33}
  & \underline{14.76} & \underline{54.35} & \underline{27.15}
  & \underline{90.41} & \underline{98.84} & \underline{91.47} \\
\bottomrule
\end{tabular}
}% end resizebox
\caption{Retrieval performance at depth $k{=}3$ across six benchmarks (IE\,$@3\uparrow$, Precision\,$@3\uparrow$, Recall\,$@3\uparrow$). \textbf{Bold} = best; \underline{underline} = second-best. \colorbox{oursblue}{\strut\,\name{}\,} rows are shaded.}
\label{tab:results_at3}
\vspace{-1em}
\end{table*}

Tables~\ref{tab:results_at1}, \ref{tab:results_at3}, and \ref{tab:results_at5} report retrieval performance at cutoffs $k{=}1$, $k{=}3$, and $k{=}5$, respectively, across all six benchmarks. As expected, both precision and recall increase monotonically with $k$ for all methods, since retrieving more documents provides greater coverage of relevant passages. The relative ordering of methods remains consistent across all cutoffs: \name{} outperforms all non-oracle baselines at every depth while staying within a narrow margin of the LLM\,+\,10\,Topics oracle, which relies on an LLM at retrieval time. This consistency demonstrates that the gains from topic-guided retrieval are not specific to any particular cutoff, but reflect a robust improvement in retrieval quality across the full range of evaluation settings reported here.

\begin{table*}[t]
\centering
\setlength{\tabcolsep}{4.5pt}
\renewcommand{\arraystretch}{1.1}

\resizebox{\textwidth}{!}{%
\begin{tabular}{lccccccccc}
\toprule
\multirow{2}{*}{\textbf{Method}}
& \multicolumn{3}{c}{\textbf{Dragonball}} & \multicolumn{3}{c}{\textbf{HotpotQA}} & \multicolumn{3}{c}{\textbf{SQuAD}} \\
\cmidrule(lr){2-4}\cmidrule(lr){5-7}\cmidrule(lr){8-10}
  & \textit{IE$@5\uparrow$} & \textit{Prec.$@5\uparrow$} & \textit{Rec.$@5\uparrow$}  & \textit{IE$@5\uparrow$} & \textit{Prec.$@5\uparrow$} & \textit{Rec.$@5\uparrow$}  & \textit{IE$@5\uparrow$} & \textit{Prec.$@5\uparrow$} & \textit{Rec.$@5\uparrow$} \\
\midrule
RAPTOR
  & 6.16 & 43.15 & 14.28
  & 12.09 & 64.63 & 18.70
  & 9.09 & 36.52 & 24.88 \\
Meta-Chunking-MSP
  & 6.76 & 43.95 & 15.38
  & 15.92 & \underline{69.30} & 22.97
  & 17.44 & 45.72 & 38.15 \\
Meta-Chunking-PPL
  & 9.07 & 46.55 & 19.48
  & 18.54 & \textbf{70.23} & 26.40
  & 16.90 & 45.12 & 37.45 \\
DenseXRetrieval
  & 0.02 & 8.15 & 0.20
  & 5.79 & 48.17 & 12.03
  & 8.23 & 34.92 & 23.58 \\
SAKI-RAG
  & 21.90 & 75.12 & 29.15
  & 21.23 & 60.60 & 35.03
  & 76.52 & 92.55 & 82.68 \\
\hline
LLM
  & 24.93 & 80.28 & 31.05
  & 23.42 & 60.83 & 38.50
  & 82.52 & 95.38 & 86.52 \\
LLM + 10 Topics
  & \textbf{34.58} & \textbf{91.18} & \textbf{37.92}
  & \textbf{30.69} & 64.33 & \textbf{47.70}
  & \textbf{92.86} & \textbf{99.58} & \textbf{93.25} \\
\hline
\rowcolor{oursblue}
MCompassRAG + 10 Topics
  & \underline{31.29} & \underline{86.55} & \underline{36.15}
  & \underline{28.02} & 61.40 & \underline{45.63}
  & \underline{91.83} & \underline{99.12} & \underline{92.65} \\
\bottomrule
\end{tabular}
}% end resizebox

\vspace{2pt}

\resizebox{\textwidth}{!}{%
\begin{tabular}{lccccccccc}
\toprule
\multirow{2}{*}{\textbf{Method}}
& \multicolumn{3}{c}{\textbf{DRBench}} & \multicolumn{3}{c}{\textbf{LegalBench-RAG}} & \multicolumn{3}{c}{\textbf{SCI-DOCS}} \\
\cmidrule(lr){2-4}\cmidrule(lr){5-7}\cmidrule(lr){8-10}
  & \textit{IE$@5\uparrow$} & \textit{Prec.$@5\uparrow$} & \textit{Rec.$@5\uparrow$}  & \textit{IE$@5\uparrow$} & \textit{Prec.$@5\uparrow$} & \textit{Rec.$@5\uparrow$}  & \textit{IE$@5\uparrow$} & \textit{Prec.$@5\uparrow$} & \textit{Rec.$@5\uparrow$} \\
\midrule
RAPTOR
  & 4.67 & 37.15 & 12.57
  & 4.30 & 35.98 & 11.95
  & 71.72 & 85.89 & 83.50 \\
Meta-Chunking-MSP
  & 6.76 & 40.51 & 16.68
  & 5.91 & 39.85 & 14.82
  & 73.85 & 86.66 & 85.22 \\
Meta-Chunking-PPL
  & 8.62 & 41.95 & 20.55
  & 7.15 & 41.28 & 17.32
  & 1.39 & 20.73 & 6.70 \\
DenseXRetrieval
  & 2.92 & 29.75 & 9.80
  & 2.55 & 28.68 & 8.88
  & 64.15 & 82.45 & 77.80 \\
SAKI-RAG
  & 21.74 & 66.67 & 32.61
  & 11.52 & 50.05 & 23.02
  & 83.39 & 95.39 & 87.42 \\
\hline
LLM
  & 26.68 & 72.81 & 36.65
  & 14.01 & 54.15 & 25.88
  & 88.98 & 98.22 & 90.59 \\
LLM + 10 Topics
  & \textbf{41.74} & \textbf{87.55} & \textbf{47.67}
  & \textbf{21.15} & \textbf{63.22} & \textbf{33.45}
  & \textbf{95.62} & \textbf{100.00} & \textbf{95.62} \\
\hline
\rowcolor{oursblue}
MCompassRAG + 10 Topics
  & \underline{37.80} & \underline{82.94} & \underline{45.58}
  & \underline{18.63} & \underline{58.85} & \underline{31.65}
  & \underline{95.22} & \underline{100.00} & \underline{95.22} \\
\bottomrule
\end{tabular}
}% end resizebox
\caption{Retrieval performance at depth $k{=}5$ across six benchmarks (IE\,$@5\uparrow$, Precision\,$@5\uparrow$, Recall\,$@5\uparrow$). \textbf{Bold} = best; \underline{underline} = second-best. \colorbox{oursblue}{\strut\,\name{}\,} rows are shaded.}
\label{tab:results_at5}
\vspace{-1em}
\end{table*}

%% file: sections/app_topic_granularity.tex
\section{Effect of Topic Granularity of Topic Model}
\label{app:topic_granularity}

Table~\ref{tab:granularity} reports retrieval performance as a function of the number of topics $K$ in the underlying topic model. Two consistent patterns emerge across all three benchmarks. First, \textbf{performance peaks at $K=100$ and degrades monotonically as $K$ increases beyond this point}. At very high granularities ($K=500$--$2000$), topic representations become increasingly fine-grained and sparse, making each topic centroid less representative of a coherent semantic direction. As a result, the weighted aggregation of topic centroids produces chunk and query representations that are noisier and harder to match reliably. Second, \textbf{the student--teacher gap is largest at $K=100$ and nearly vanishes at high $K$}. At $K=100$, the LLM teacher can exploit the richer and more semantically coherent per-topic structure to outperform the student, which receives only a compressed topic summary. At $K \geq 500$, both the teacher and the student suffer equally from the degraded topic quality, and their performance converges. Together, these results suggest that \textbf{a moderate topic granularity of $K=100$ strikes the best balance} between topic coherence and coverage, and we use this setting across all experiments in the main paper. This finding is complementary to the analysis in Section~\ref{sec:ablations}, which studied the effect of how many topic signals are passed to the model at inference time: here we show that the quality of those signals, determined by $K$, is equally important. Even with an optimal number of passed topics, overly fine-grained or coarse topic models will degrade retrieval quality.

\begin{table*}[t]
\centering
\resizebox{\textwidth}{!}{%
\begin{tabular}{@{}ll ccccccccc@{}}
\toprule
& & \multicolumn{3}{c}{\textbf{SCI-DOCS}} & \multicolumn{3}{c}{\textbf{LegalBench-RAG}} & \multicolumn{3}{c}{\textbf{Dragonball}} \\
\cmidrule(lr){3-5} \cmidrule(lr){6-8} \cmidrule(lr){9-11}
$\textbf{K}$ & \textbf{Method} & {Recall $\uparrow$} & {Precision $\uparrow$} & {IE $\uparrow$} & {Recall $\uparrow$} & {Precision $\uparrow$} & {IE $\uparrow$} & {Recall $\uparrow$} & {Precision $\uparrow$} & {IE $\uparrow$} \\
\midrule
\multirow{2}{*}{50}
  & MCompassRAG & 88.83 & 93.37 & 86.87 & 36.23 & 51.40 & 26.30 & 36.73 & 78.53 & 30.47 \\
  & LLM         & 92.43 & 98.13 & 91.90 & 37.70 & 55.43 & 27.83 & 38.43 & 82.60 & 32.07 \\
\midrule
\multirow{2}{*}{100}
  & MCompassRAG & 94.13 & 99.03 & 92.10 & 38.40 & 55.10 & 27.90 & 38.97 & 82.80 & 32.40 \\
  & LLM         & 98.30 & 99.63 & 98.03 & 40.10 & 59.47 & 29.70 & 40.83 & 87.43 & 34.17 \\
\midrule
\multirow{2}{*}{500}
  & MCompassRAG & 86.53 & 89.63 & 83.47 & 35.30 & 49.87 & 25.27 & 35.60 & 74.90 & 28.77 \\
  & LLM         & 87.20 & 90.40 & 84.03 & 35.57 & 50.30 & 25.43 & 35.97 & 75.37 & 28.97 \\
\midrule
\multirow{2}{*}{1000}
  & MCompassRAG & 84.80 & 87.10 & 81.17 & 34.60 & 48.47 & 24.57 & 34.63 & 72.83 & 27.80 \\
  & LLM         & 85.13 & 87.47 & 81.50 & 34.73 & 48.67 & 24.67 & 34.83 & 73.07 & 27.90 \\
\midrule
\multirow{2}{*}{2000}
  & MCompassRAG & 83.40 & 85.23 & 79.60 & 34.03 & 47.43 & 24.10 & 33.90 & 71.27 & 27.07 \\
  & LLM         & 83.57 & 85.40 & 79.83 & 34.10 & 47.53 & 24.17 & 34.00 & 71.40 & 27.13 \\
\bottomrule
\end{tabular}%
}
\caption{Effect of topic model granularity ($K$) on retrieval performance across three datasets. Results are reported for MCompassRAG and LLM-based methods.}
\label{tab:granularity}
\end{table*}

%% file: sections/app_in_domain_topic_model.tex
\section{Topic Model Domain Adaptation: Training on Target Corpus}
\label{app:in_domain_topic_model}

In the main experiments, we use a topic model trained on WikiWeb2M to provide a general-purpose set of topic centroids and document-topic vectors. While this setting tests whether \name{} can rely on a broadly trained topic model, some benchmarks contain domain-specific terminology and evidence structures that may not be fully captured by a general corpus. We therefore evaluate an in-domain variant in which the topic model is trained directly on the target corpus of each benchmark, while keeping the rest of the \name{} pipeline unchanged.

Table~\ref{tab:in_domain_topic_model} compares the default WikiWeb2M-trained topic model with target-corpus topic models on Dragonball, LegalBench-RAG, and SCI-DOCS. Training the topic model on the target corpus improves performance across all three datasets, with larger gains on LegalBench-RAG and Dragonball, where domain-specific terminology, entities, and narrative structure are especially important. However, the gains are moderate rather than dramatic, showing that \name{} does not require retraining the topic model for every new corpus. This is important for practical deployment: a general-purpose topic model can already provide useful metadata guidance, while in-domain topic modeling can be used as an optional enhancement when sufficient target-corpus data and training budget are available.

\begin{table*}[t]
    \centering
    \setlength{\tabcolsep}{4.0pt}
    \renewcommand{\arraystretch}{1.12}
    \resizebox{0.85\textwidth}{!}{%
    \begin{tabular}{lccccccccc}
    \toprule
    \multirow{2}{*}{\textbf{Topic Model Training Corpus}}
      & \multicolumn{3}{c}{\textbf{Dragonball}}
      & \multicolumn{3}{c}{\textbf{LegalBench-RAG}}
      & \multicolumn{3}{c}{\textbf{SCI-DOCS}} \\
    \cmidrule(lr){2-4}\cmidrule(lr){5-7}\cmidrule(lr){8-10}
      & \textit{IE$\uparrow$} & \textit{Prec.$\uparrow$} & \textit{Rec.$\uparrow$}
      & \textit{IE$\uparrow$} & \textit{Prec.$\uparrow$} & \textit{Rec.$\uparrow$}
      & \textit{IE$\uparrow$} & \textit{Prec.$\uparrow$} & \textit{Rec.$\uparrow$} \\
    \midrule
    WikiWeb2M
      & 38.97 & 82.80 & 32.40
      & 38.40 & 55.10 & 27.90
      & 94.13 & 99.03 & 92.10 \\
      
    Target Corpus
      & \textbf{39.26} & \textbf{83.71} & \textbf{32.83}
      & \textbf{40.18} & \textbf{57.36} & \textbf{29.64}
      & \textbf{94.82} & \textbf{99.21} & \textbf{92.86} \\
    \bottomrule
    \end{tabular}
    }
    \caption{
    Effect of training the topic model on the target corpus. Results report IE$\uparrow$, Precision$\uparrow$, and Recall$\uparrow$, averaged over retrieval cutoffs $k{=}1,3,5$. 
    The WikiWeb2M row corresponds to the main \name{} configuration, while the Target Corpus row trains the topic model on the corresponding benchmark corpus before running the same retrieval pipeline.
    }
    \label{tab:in_domain_topic_model}
\end{table*}

%% file: sections/app_qualitative_analysis.tex
\section{Qualitative Analysis}
\label{app:qualitative}

\begin{figure*}[t]
  \centering
  \includegraphics[width=\linewidth]{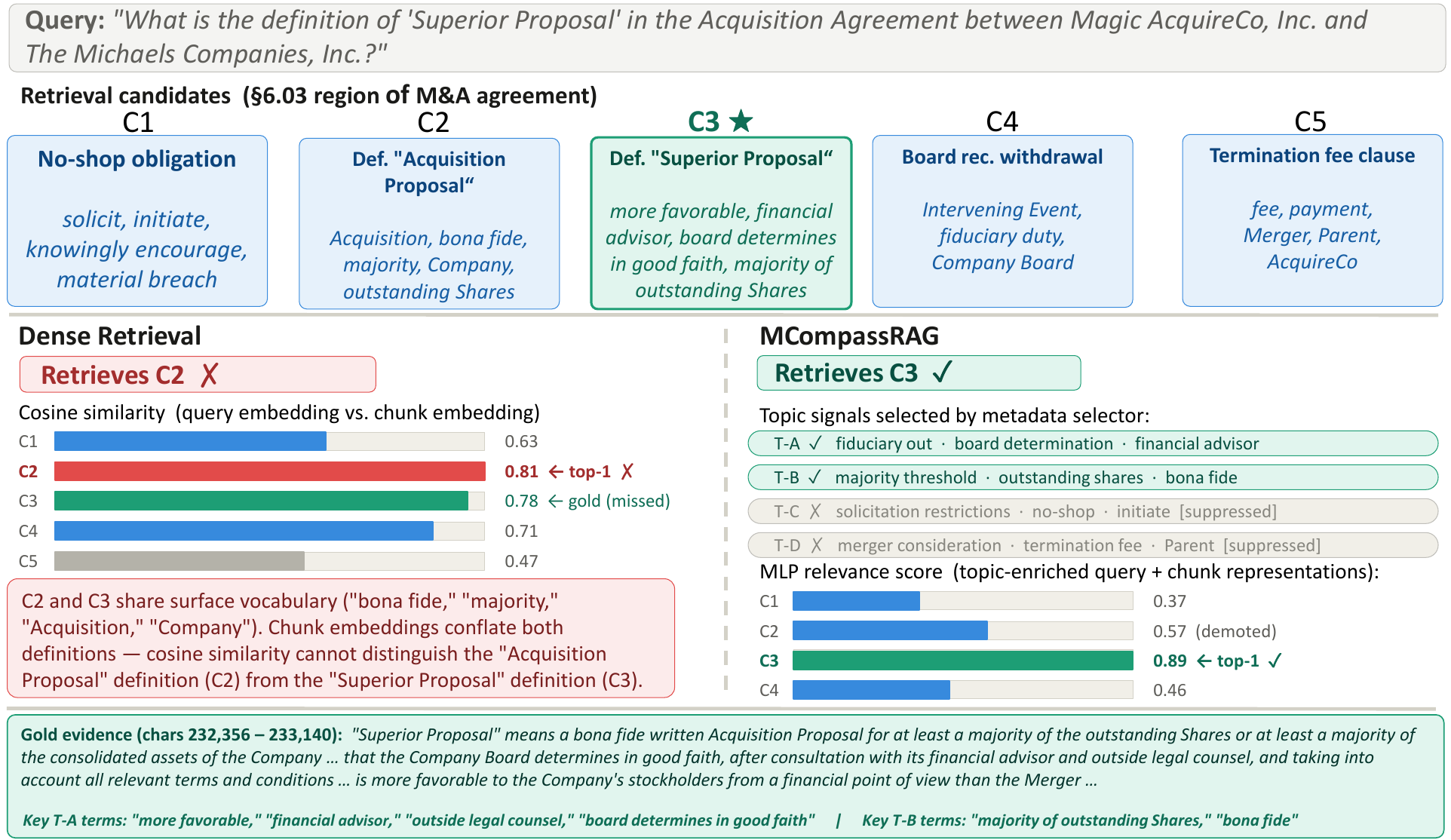}
  \caption{Qualitative retrieval comparison on LegalBench-RAG for a query about the definition of \textit{Superior Proposal} in an M\&A acquisition agreement. \textbf{Top}: five retrieval candidates from the \S6.03 region; the gold chunk (C3, teal border) competes against four topically adjacent clauses sharing substantial surface vocabulary. \textbf{Bottom left}: dense retrieval ranks C2 (\textit{Acquisition Proposal} definition) above C3 due to overlapping tokens, missing the gold chunk at rank 1. \textbf{Bottom right}: \name{} activates topic signals T-A (fiduciary out / board determination) and T-B (majority threshold), suppresses T-C and T-D, and promotes C3 to rank 1 via the MLP scorer (0.89 vs.\ 0.57 for C2).}
  \label{fig:qualitative}
\end{figure*}

We present two qualitative examples to illustrate how \name{} resolves retrieval failures that dense similarity cannot handle: a definitional ambiguity case from LegalBench-RAG and an embedding-space analysis from Dragonball Finance.

\subsection{LegalBench-RAG: definitional ambiguity in M\&A agreements.} 
Figure~\ref{fig:qualitative} illustrates a concrete retrieval example from LegalBench-RAG that exposes the core failure mode of dense retrieval and how \name{} resolves it. The query asks for the definition of \textit{Superior Proposal} in an M\&A acquisition agreement whose \S6.03 region contains several topically adjacent clauses: a no-shop obligation (C1), the definition of \textit{Acquisition Proposal} (C2), a board recommendation withdrawal clause (C4), and a termination fee clause (C5). Dense retrieval assigns the highest cosine similarity to C2 (0.81) rather than the gold chunk C3 (0.78), ranking the wrong definition first. The failure arises because C2 and C3 share substantial surface vocabulary (``bona fide,'' ``majority,'' ``Acquisition,'' ``outstanding Shares''), causing their embeddings to occupy nearby positions in the retriever space. Cosine similarity cannot identify which latent topic of a chunk matches the query, nor distinguish a clause that \textit{defines} what counts as an acquisition proposal from one that \textit{evaluates} whether a proposal is superior.

\name{} recovers the gold chunk by activating two topic signals identified by the metadata selector as compatible with the query embedding: T-A, capturing the fiduciary-out and board determination frame (``more favorable,'' ``financial advisor,'' ``board determines in good faith''), and T-B, capturing the majority threshold frame (``majority of outstanding Shares,'' ``bona fide written Acquisition Proposal''). The selector simultaneously suppresses signals associated with solicitation restrictions (T-C) and merger consideration (T-D), which are prominent in the neighboring chunks but orthogonal to the query's information need. The abstraction module pools T-A and T-B into a compact query-side topic vector aligned with C3's own topic representation, and the MLP classifier assigns C3 a relevance score of 0.89 versus 0.57 for C2, promoting the correct definition to rank 1 without any inference-time LLM call. This disambiguation was learned through the teacher--student asymmetry in Section~\ref{sec:training}: the LLM teacher, given the expanded query framing the information need in terms of board determination and financial-advisor consultation, labels C3 as relevant and C2 as not, training the student to recover the same judgment through topic metadata alone.

\subsection{Dragonball Finance: topic-guided separation in embedding space.}
\begin{figure*}[t]
  \centering
  \includegraphics[width=\linewidth]{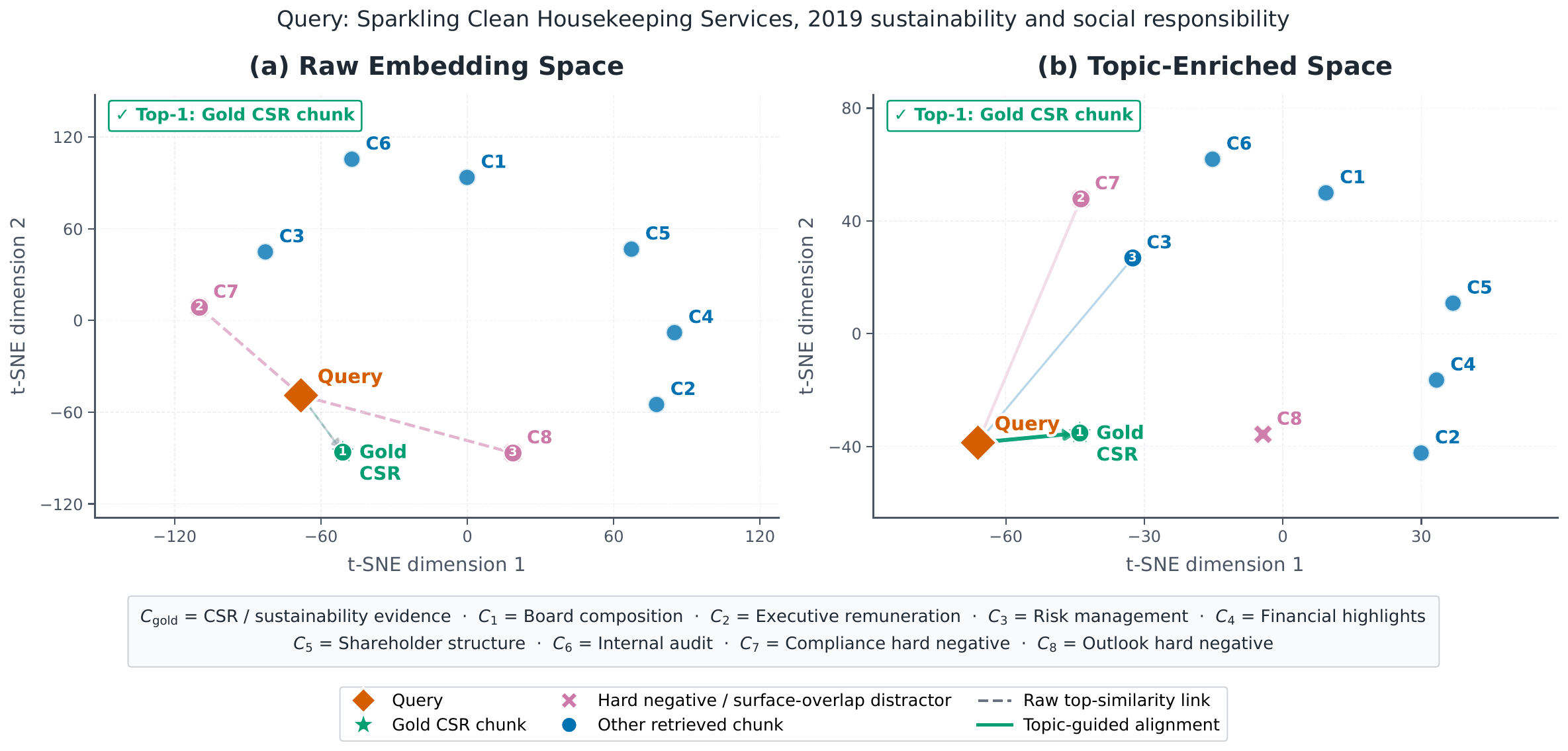}
  \caption{t-SNE visualization of chunk embeddings for a Dragonball Finance query on Sparkling Clean Housekeeping Services' 2019 sustainability efforts. Chunks cover eight aspects of the corporate governance report: board composition (C1), executive remuneration (C2), risk management (C3), financial highlights (C4), shareholde structure (C5), internal audit (C6), compliance and anti-corruption (C7), and strategic outlook (C8); C7 and C8 are surface-overlap distractors that share phrases with the gold CSR chunk. \textbf{(a) Raw embedding space}: the query and gold chunk are proximate but several hard negatives occupy the same neighbourhood. \textbf{(b) Topic-enriched space}: topic enrichment tightens the query--gold alignment while pushing all hard negatives, including the surface-overlap distractors C7 and C8, away from the query.}
  \label{fig:tsne}
\end{figure*}

Figure~\ref{fig:tsne} visualizes the effect of topic enrichment on a Dragonball Finance example in which the query asks for a summary of Sparkling Clean Housekeeping Services' sustainability and social responsibility efforts in 2019. The eight retrieval candidates span the full thematic range of the company's corporate governance report: board composition (C1), executive remuneration (C2), risk management (C3), financial highlights (C4), shareholder structure (C5), internal audit (C6), and two surface-overlap distractors whose vocabulary partially overlaps with the gold chunk: a compliance and anti-corruption clause (C7, which shares the phrase ``corporate citizenship'') and a strategic outlook statement (C8, which shares ``long-term value creation'').

In the raw embedding space (Figure~\ref{fig:tsne}a), the query and the gold CSR chunk are already relatively proximate, yet several hard negatives remain in the same neighbourhood, reflecting the broad semantic overlap that coarse governance-report language introduces. After topic enrichment (Figure~\ref{fig:tsne}b), the query--gold alignment tightens substantially: the metadata selector activates the CSR topic centroid for the query and the gold chunk's own topic distribution loads on the same signal, pulling the two representations into close alignment while the hard negatives, whose dominant topic vectors correspond to governance, finance, and risk, drift away. The surface-overlap distractors C7 and C8 are particularly informative: despite sharing specific phrases with the gold chunk, their topic distributions do not load on the CSR centroid and therefore receive lower relevance scores from the MLP classifier, confirming that \name{}'s disambiguation operates at the level of latent topic structure rather than lexical overlap.

